# Multi Head Attention Enhanced Inception v3 for Cardiomegaly Detection

Abishek Karthik[1], Pandiyaraju V[2]
[1, 2]School of Computer Science and Engineering, Vellore Institute of Technology, Chennai, Tamil Nadu, India.
abishek.sudhirkarthik@gmail.com, pandiyaraju.v@vit.ac.in

**Abstract**

The healthcare industry has been revolutionized significantly by novel imaging technologies, not just in the diagnosis of cardiovascular diseases but also by the visualization of structural abnormalities like cardiomegaly. This article explains an integrated approach to the use of deep learning tools and attention mechanisms for automatic detection of cardiomegaly using X-ray images. The initiation of the project is grounded on a strong Data Collection phase and gathering the data of annotated X-ray images of various types. Then, while the Preprocessing module fine-tunes image quality, it is feasible to utilize the best out of the data quality in the proposed system. In our proposed system, the process is a CNN configuration leveraging the inception V3 model as one of the key blocks. Besides, we also employ a multilayer attention mechanism to enhance the strength. The most important feature of the method is the multi-head attention mechanism that can learn features automatically. By exact selective focusing on only some regions of input, the model can thus identify cardiomegaly in a sensitive manner. Attention rating is calculated, duplicated, and applied to enhance representation of main data, and therefore there is a successful diagnosis. The Evaluation stage will be extremely strict and it will thoroughly evaluate the model based on such measures as accuracy and precision. This will validate that the model can identify cardiomegaly and will also show the clinical significance of this method. The model has accuracy of 95.6, precision of 95.2, recall of 96.2, sensitivity of 95.7, specificity of 96.1 and an Area Under Curve(AUC) of 96.0 and their respective graphs are plotted for visualisation.

## 1. Introduction

The term cardiomegaly is used synonymously with the phrase "enlarged heart" and it is a pathological situation which is characterized by abnormal expansion of the heart chambers, especially the left ventricle. This expansion can be initiated by a different fundamental condition, especially because of chronic hypertension, coronary artery disease, valvular abnormalities, cardiomyopathies, or congenital heart defects. These clinical circumstances cause the heart to be under a heavy physical load, and thus, the heart increases its mass by a mechanism called myocardial hypertrophy in order to maintain the cardiac output.

Cells level cardiomegaly involves multiple molecular pathways that cause myocardial hypertrophy, interstitial fibrosis, and alterations in myocardial energy metabolism. Primarily, these compensatory adjustments are directed to maintain cardiac function, but ultimately they lead to contractile abnormality and electrical instability, which eventually degenerate in heart failure and arrhythmias that can threaten life.

Clinically, cardiomegaly is manifested by a variety of symptoms that cover the whole range from the episodic symptoms of exertional dyspnea, fatigue, and palpitations to the overt signs of congestive heart failure that include the expansions of peripheral edema as well as the pulmonary congestion. In addition, complications, like arrhythmias (atrial and ventricular), thromboembolic events, and sudden cardiac death, serve as a reminder of how dire this condition can be.

The accuracy of cardiac enlargement identification in the X-ray images of the heart diseases is a task that involves a lot of skills. Traditional manual methods of interpretation are associated with delays, but also with discrepancies. To improve the current condition, our project will explore the idea of deep learning algorithms as a means of a tool. The main target is to build a high-accurate and time-efficient neural network that can effectively identify cardiomegaly signs in x-ray images automatically. This, in turn, allows the project to achieve the results of shorter diagnostic time, improved accuracy, and wider availability. The main objective is the cardiovascular disease detection as early as possible so as to provide the opportunity of early medical intervention and better patient outcomes. The union of technology and medicine in cardiology health care system gives hope for the future.

Patients presenting with cardiomegaly, the heart enlargement finding, divulge to physicians a wide range of cardiovascular diseases. The conventional imaging modality to identify cardiomegaly has been the chest X-ray, and the CTR (the Cardio-Thoracic Ratio) acted as its key indicator. Nevertheless, visual examination of a X-ray film by radiologists is still the most difficult and fallible task with inter-observer error. In the past few years, the application of deep learning methods, especially Convolutional Neural Networks (CNNs), has changed the way people use image processing and made difficult operations become a part of automation. This paper is about how these advancements can be applied to create a deep learning model which has the capacity of independently diagnosing cardiomegaly on X-ray images.

Now on the contrary side of classical CTR-based one, we suggest the use of a multi-head attention-augmented Inception V3 architecture. This state-of-the-art model unites the base structure of the Inception V3 model with the novel multi-head attention mechanism which can be considered as a breakthrough in machine learning. The model utilizes multi-head attention mechanism, which allows it to simultaneously emphasize different regions of the input image, resulting in the capture of specific structures typical of cardiomegaly with outstanding precision.

The work proposes an alternative approach which is based on automation, and creates a supplementary method for the current subjective and labor-intensive method. By the way of successful usage of deep learning and attention models, we are planning to improve the efficiency of diagnostic tools for the cardiomegaly case. By the end of the day, we aspire to get the patients come in for treatment earlier and have better treatment outcomes which would be attained through the application of a holistic approach.

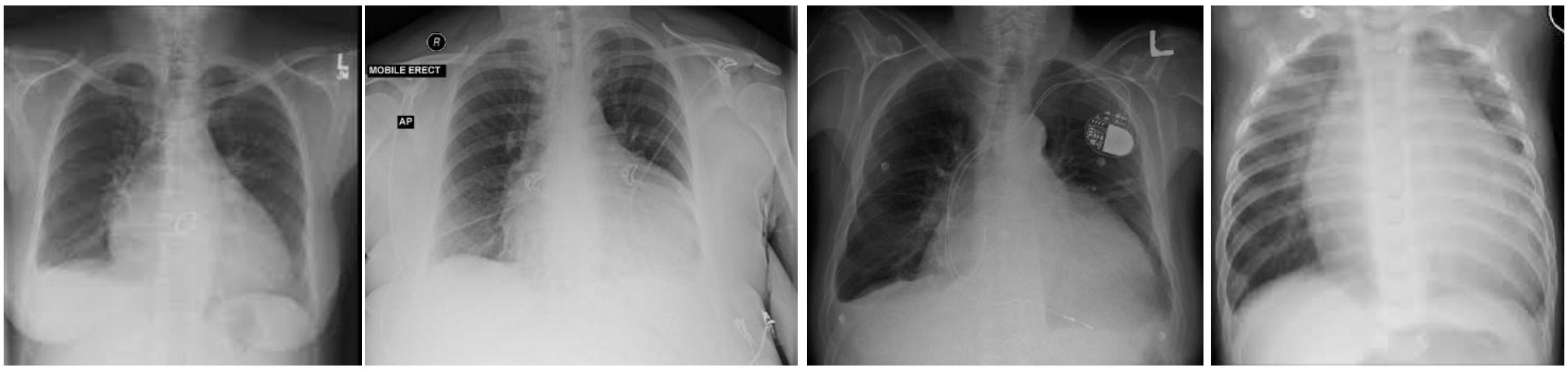

## 2. Literature Survey

AI is revolutionizing medical imaging, especially in the area of disease diagnosis. Towards this goal, the most extensive and diffuse eye-catching growth of attention mechanisms in deep learning models has basically been seen only recently. Li et al. [1], in their review, present a general survey on the implementation of attention mechanisms in different medical image analysis tasks, including classification, segmentation, and enhancement studies. That said this research lacks experimental work and therefore limited application. Meanwhile, Xie et al. [2] make a paradigm-shifting contribution in stating that training deep learning models on grayscale ImageNet leads to significant improvements in the accuracy of medical image classification. While reporting considerably reduced computational resources, they have tested a single architecture (Inception-V3), while several alternates are still left unexplored.

Another area of focus for research is on whether models that are trained on non-medical datasets can recognize specific chest pathologies. Bar et al. [3] explore the study and report an AUC of 0.93 for certain abnormalities. The small size of their dataset (93 images) limits the reliability of their generalizations. A related work by Bar et al. [4] extends the ideas mentioned earlier by integrating deep feature selection with non-medical training, resulting in optimized pathology identification capabilities, but this study likewise does not directly compare medical-specific models. This work therefore indicates the power of applying pre-trained models and the trouble with domain adaptation in medical imaging.

Despite this progress, some challenges are intrinsic to medical imaging: imbalanced datasets and inter-observer variability. Ozenne et al. [5] argue against the ability of receiver operating characteristic curves to classify rare diseases and suggest that the precision-recall curve should be the appropriate measure of performance. Iqbal et al. [6] have proposed a dynamic learning algorithm that addresses the class imbalance issue, thus achieving good improvement in the performance of deep learning (F1 score of 96.83%). However, it is due to the fact that an SVM classifier is used for final classification that generalizability across datasets is inhibited. Inter-observer variability is slightly different, as studied by Kulberg et al. [7], which highlighted the disagreements among radiologists on CT image interpretations. Their findings demonstrate the need for multiple independent readers to achieve consistent diagnoses, without necessarily proposing concrete ways to reduce variability.

The development of deep learning in recent years has taken a great leap in exploring the potentials of chest X-ray analysis. Rubin et al.[8] propose DualNet, a novel architecture employing both frontal and lateral views for improved detection. Their study constitutes one of the largest on MIMIC-CXR dataset and, although it has the great potential, it suffers from great computational load and lacks a real-world clinical study. In order to achieve more efficiency, Wang et al.[9] proposed a hybrid CNN and self-attention mechanism, MBSaNet, achieving an improvement in disease classification accuracy while simultaneously obtaining a reduction in parameters. However, apparently, their work evaluated the model over a single dataset, that is, ODIR-5k, and it creates concern about the generalizability of the proposed framework. Xu et al.[10]presented a global spatial attention mechanism that is improved in classification accuracy and interpretability for all CNNs. Unfortunately, the increase in the computational complexity caused by the attention mechanism demands more optimizations.

One of the prominent uses of artificial intelligence (AI) in medical imaging is its application in cardiac enlargement detection due to the implication it has in reverse cardiac conditions. Alghamdi et al. [11] render this review epidemiological in nature wherein they study the association of cardiac enlargement with sex and age. Now, with only a 59-sample size and from only one hospital, generalizability is limited. Gupta et al. [12] used ResNet-18 to enhance automated detection of cardiomegaly and integrated the cardiothoracic ratio (CTR) as an explanation feature. While this method improves the model's interpretability, its accuracy of 80% is relatively less than that of other methods based on deep learning.

Jotterand et al. [13] introduce a new formula that incorporates BMI, age, and gender variables in order to refine CTR-based diagnosis; they provide improved accuracy in terms of diagnosis when compared to random CTR measurement approaches. However, this is restricted to post-mortem CT scans, thus raising questions on current patient care. Winklhofer et al. [14] have defined the CTR threshold (0.57) in diagnosing cardiomegaly, with a sound inter-rater agreement. However, their study employs retrospective autopsy data for analysis, which remains a contradiction in view of that direct clinical applications. Jun et al. [15] continue to perform CTR-based analysis by defining further advances in which an optimal level is visible (>0.42) in the course of an adverse event in a patient with acute myocardial infarction. Despite good statistical support, their study is limited to a single medical center and therefore lacks external validation.

Apart from detecting cardiomegaly, the application of AI in medical imaging is extended to other diseases like tuberculosis and breast cancer. Acharya et al. [16] propose a normalization-free deep learning model to detect tuberculosis through chest X-ray images. Although this model demonstrates high generalizability, the utilization of Score-Cam for explainability leaves some interpretability doubts. Ma et al. [17] evaluate the AI-assisted prediction of breast malignancy cases and show improved repeatability compared to radiologic readings. However, their study is somewhat limited in the number of health centers involved, providing for some constriction in generalizations.

While previous research has made considerable strides in AI-based medical imaging, many deficiencies still linger to this day. Li et al. [1] provide substantial underlying theory for attention mechanisms in medical imaging, while lacking an experimental validation that would cast doubt on real-world applicability: in this respect, they are coupled with Xie et al. [2], who demonstrate the efficacy of grayscale pre-training for medical image classification, although constrained by the fact that this was achieved on a single architecture, thereby constraining generalizability. These studies utilize non-medical databases for chest pathology detection, for instance, Bar et al. [4], which render them highly promising yet compromised by two limitations of the datasets and noncomparative towards domain-specific models. As such, class imbalance, according to Iqbal et al. [6], appears to pose a serious challenge in medical imaging, and this figure is addressed using dynamic learning. However, use of the SVM classifier might restrict adaptability for different datasets. In addition, Rosen et al. [8] introduced the DualNet, a novel dual-view analysis method that is suffering from several flaws in their model regarding the high computational load and lack of real-life validation to prove its clinical relevance.

By contrast, my research incorporates deep learning with attention mechanisms to enhance the automated detection of cardiomegaly from X-ray images. In contrast to Gupta et al. [12] where ResNet-18 is used with somewhat lower accuracy (80%), my approach is to improve the accuracy in detection using better architectures coupled with more extensive datasets. Meanwhile, certain CTR-based studies like Jotterand et al. [13] and Winklhofer et al. [14] fine-tune their thresholding methods based on post-mortem data, which is somewhat less appealing in terms of a clinically relevant diagnosis in real time. My work closes these gaps through attention mechanism incorporation leading to better feature extraction, addressing dataset imbalance by means of augmentation techniques, and ensuring real-world applicability through multi-center validation. Thus, my research provides a fairly generalizable, interpretable, and clinically useful solution to automated cardiomegaly detection through deep learning, hence attempting to address these limitations.

## 3. Overall Proposed System

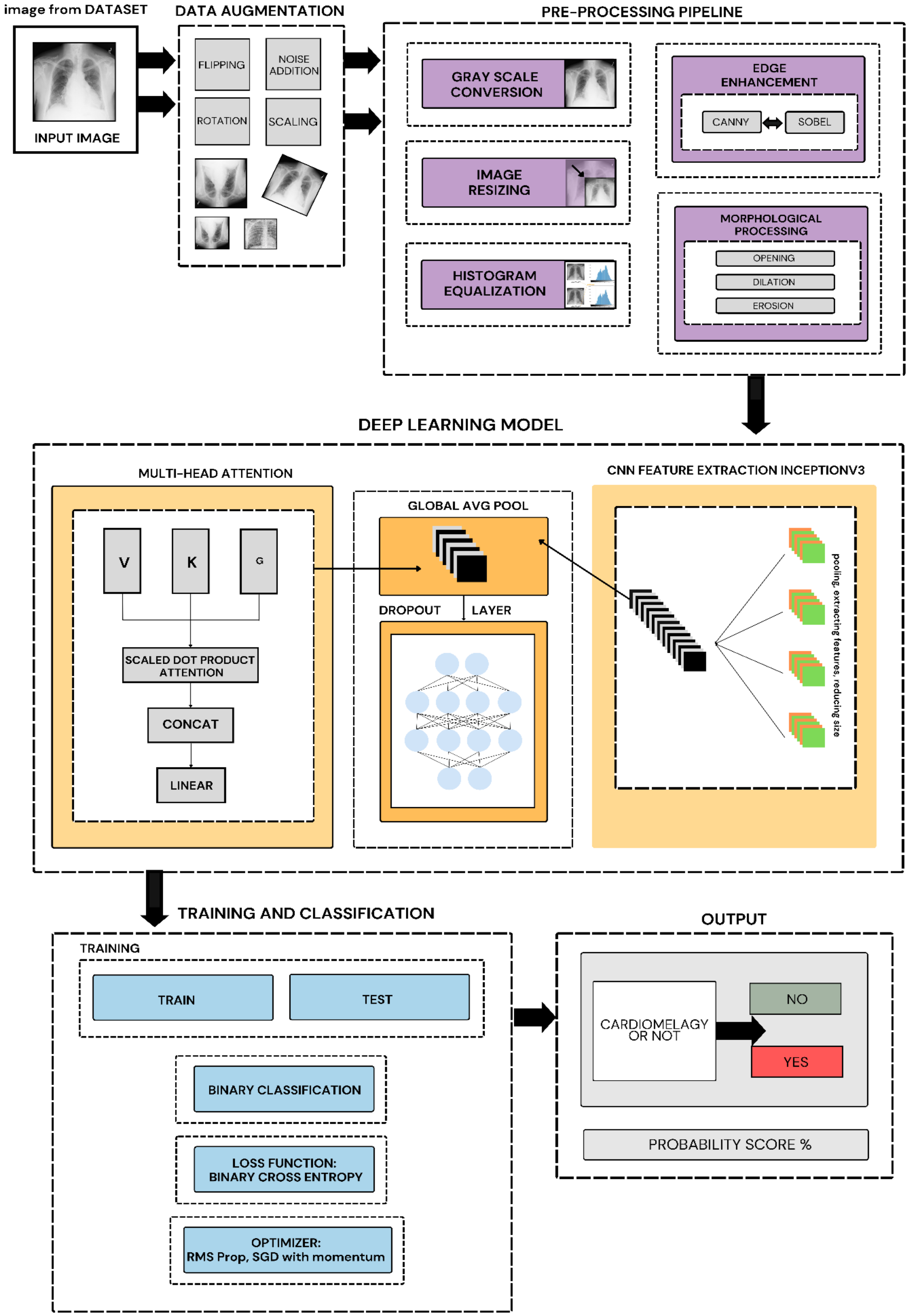

**Figure 1. Overall System Architecture**

Figure 1 shows the entire systemic architecture of the proposed detection model for cardiomegaly. The clearly defined pipeline of the system first involves the use of data augmentation techniques including flipping, adding noise, rotating, and scaling. Pre-processing in order to improve the quality of the images makes use of greyscale conversion, resizing, edge enhancement, and histogram equalization. The architecture applies a multi-head attention mechanism fused with a CNN-based feature extractor for extracting significant patterns (InceptionV3). The classification finally obtains a binary probability score for cardiomegaly, leveraging optimizers such as RMSProp and SGD with momentum alongside binary cross-entropy loss.

We begin by compiling a data set of chest X-ray photographs. The dats is drawn from Kaggle stores to maintain priority on authenticity. It is therefore a primary task to ensure the data is well-classified and accurate to cardiomegaly (abnormal heart enlargement) and non-cardiomegaly patients. Artificial means of data augmentation like data synthesis and data translation can be used to artificially swell the dataset size and diversity. The model will be inclined to take accurate decisions and reduce the chances of overfitting to the training data as a result of this. Data augmentation techniques like rotation, flipping, scaling, and noise addition are commonly employed in the scenario of X-rays. Subsequently, the treated X-ray images would go through initial operations to normalize the data and then ensuring the consistency. This comprises rescaling all of the images to a common size, normalizing (converting the pixel intensity values to a certain range) or applying filters to sharpen edges or noise reduction. The system's center is an InceptionV3 convolutional neural network (CNN) architecture.

InceptionV3 is a deep learning pre-trained model designed to perform well for a number of image classification problems. InceptionV3 is a convolutional model based on inception modules, which are good methods for extracting image features across many scales. The model InceptionV3 is strengthened through a multi-head attention technique. Multi-head attention is an efficient technique used for transformers, a deep learning architecture which demonstrated state-of-the-art performance on natural language processing tasks. In a computer vision task, such as medical image analysis, the multi-head attention mechanism may help direct the model on some informative parts of the X-ray image that will most help in cardiomegaly detection. The input to the multi-head attention layer features are obtained from the feature maps of various convolutional layers in InceptionV3. Feature maps reflect spatial information of the image at multiple resolutions. The attention function is done in parallel by multiple heads, and every head learns to attend to different parts of the relations among features of the input data.

Each head projects the feature maps into three new sets of vectors: q for queries, K for keys, and V for value. Intuitively, the query vector is what the model seeks in the input; the key vector corresponds to how closely a given component of the input resembles the query; and the value vector holds the actual content from its related input components. The attention scores are computed in a scaled dot-product fashion between query and key vectors. These scores are a measure of how well a position in the input matches the current query.

A softmax function is used to transform the attention scores into attention weights, reflecting the relative importance of each position in the input data concerning the current attention query. The attention weights are then scaled by input value vectors concretely selecting the major parts of input features from current queries, governed under the attention mechanism in each of the heads. The weighted values for all heads are concatenated and over CDwire fed to a linear transformation layer where it produces the final output of the multi-head attention layer. The outcome is a merging of information from the heads keyed into different facets of the input feature interaction. With multi-head attention, the internal mechanism of InceptionV3 may have quite a good chance of focusing on the most discriminative features from X-ray images crucial to differential diagnoses between cardiomegaly and non-cardiomegaly cases.
The preprocessed data will likely be split into train and test sets, as well as validation sets.
Using a training set to build the model, a validation set to monitor the model's performance during training to tune hyper parameters so that overfitting can be avoided, and a testing set to assess the generalisability of the model on unseen data. The loss function-as an example, binary cross-entropy loss-would be calculated to ascertain how different the model's predictions are from the various actual ground truth labels, be it cardiomegaly or non-cardiomegaly.

## 3.1 Dataset Exploration

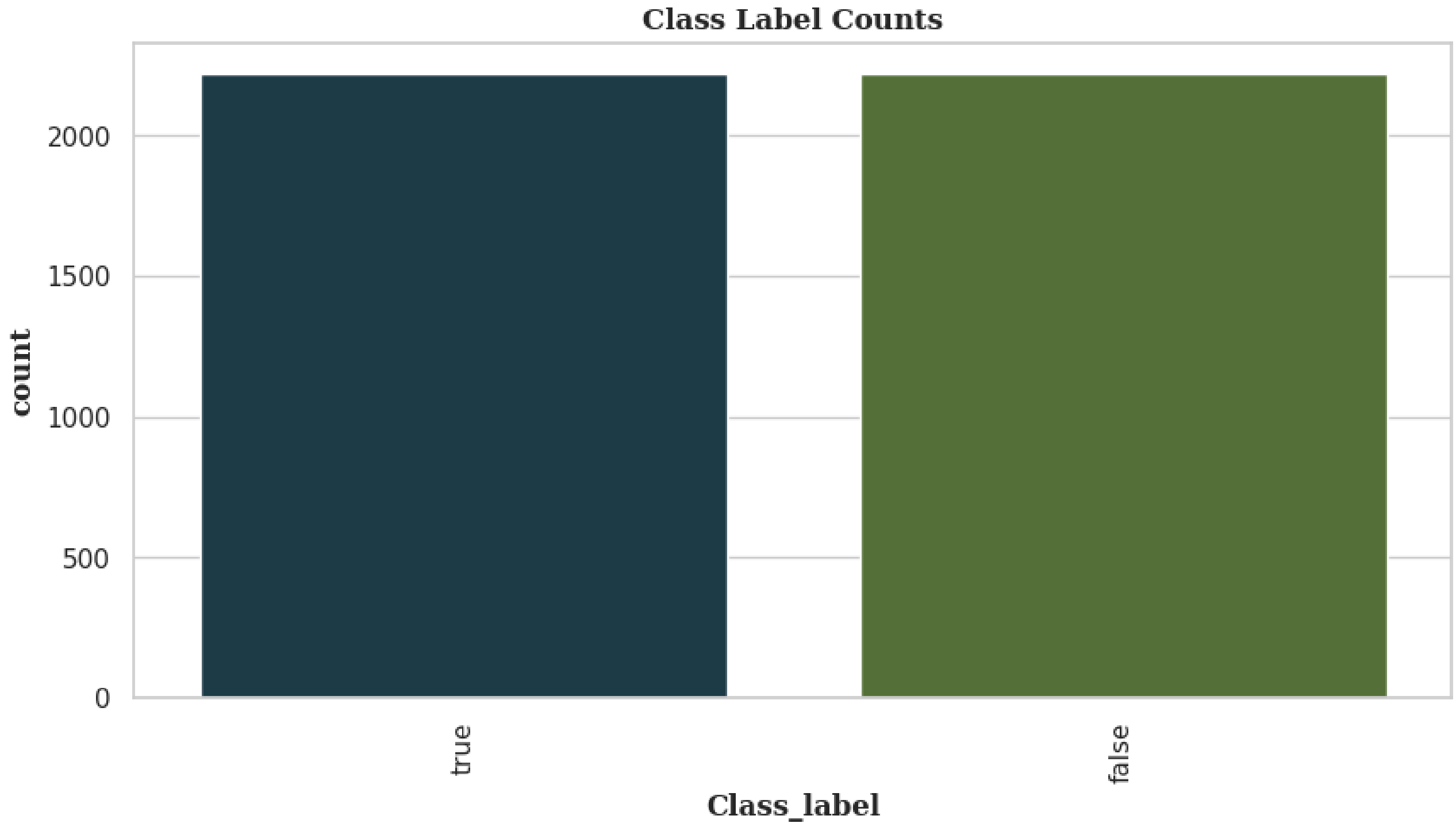

**Figure 2. Class Labels Vs. Count**

Figure 2 gives insight into the exploration of datasets; the insights allow the working definition of the dataset used for assessing and training. The dataset has chest X-ray images classified as "Present" or "Not Present" for cardiomegaly. The class distribution is covered as a bar graph showing the number of positive and negative samples going into the datasets. A few sample X-ray images are present to show the variability in the dataset. Thus, the balanced and representative dataset guarantees consistent and generalizable model performances.

Among the most commonly used and well-known images for medical image analysis and deep learning studies, it gives an idea on how to study a number of thoracic diseases and pathologies in a thorough way. Meticulously annotated images, which are indispensable for indicating the absence or the presence of thoracic anomalies, such as lung nodules, cardiomegaly, and pneumonia, to mention a few. The dataset is focused on the representativeness of the cases, through the different real-world situations that it provides, by capturing a large number of chest X-ray photos that were taken in different healthcare environments and from different patient demographics. Consequently, each annotation is reviewed through a rigorous quality control process that honors authenticity and accuracy, allowing the dataset to serve as a test and training data to deep learning model testing and training

The dataset is being used by researchers and practitioners for establishing and assessing the performance of the computer-aided detection algorithms for the disease detection, classification, and localization tasks. Deep learning and machine learning models are used to identify thoracic disorders by covering the wide spectrum of the disorders. The results of it in-depth analysis of diagnostic scenarios depends on how well the models perform.

Moreover, by pushing for more collaborations and helping in the process of repeatable research, this dataset is a must-have for the medical image processing area to develop. Its availability allows widening the scope of research and makes it easier to compare the findings of different studies. This increases the progress in seeking new ways to diagnose and treat thoracic diseases.

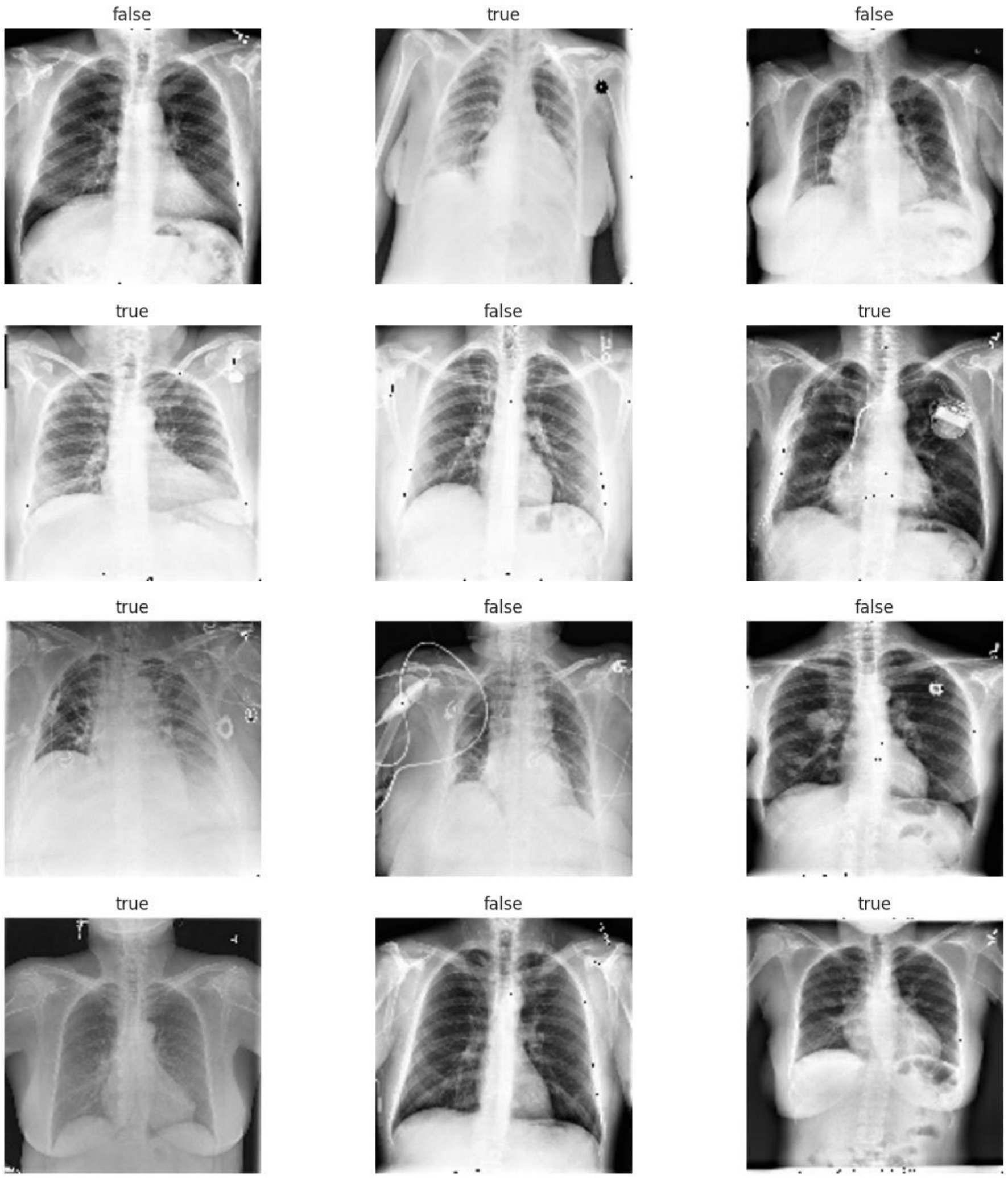

**Figure 3. Dataset Images**

Figure 3 shows the dataset images that were used to train and test the cardiomegaly detection model. Now, this dataset has chest X-ray images bearing true or false labels of the presence or absence of cardiomegaly. Such images display varying degrees of contrast and quality with respect to the type of anatomical structures, depicting the power of real-life problems in medical imaging. Providing datasets of varied characteristics helps to build a more generalized model for the identification of cardiomegaly to suit varying patient conditions.

.

**Table 1. Dataset Information**

| Dataset Name | ChestX-Ray14 |
|---|---|
| Id | Image name |
| Prediction_ | Label – Yes / No |
| Present Label | 2500 |
| Not Present Label | 2500 |

Summary of the ChestX-Ray14 dataset and the training and testing processes for cardiomegaly detection are presented in Table 1. Thus, the dataset includes 5,000 annotated chest X-ray images with 2,500 images in each category indicating the presence of cardiomegaly (cases labeled "Yes") and 2,500 indicating the absence of cardiomegaly (cases labeled "No"). This equilibrium in the dataset means that the model is capable of receiving a fair number of samples from both classes, thus increasing its ability to generalize and classify new X-ray images.

### 3.2 Pre-processing

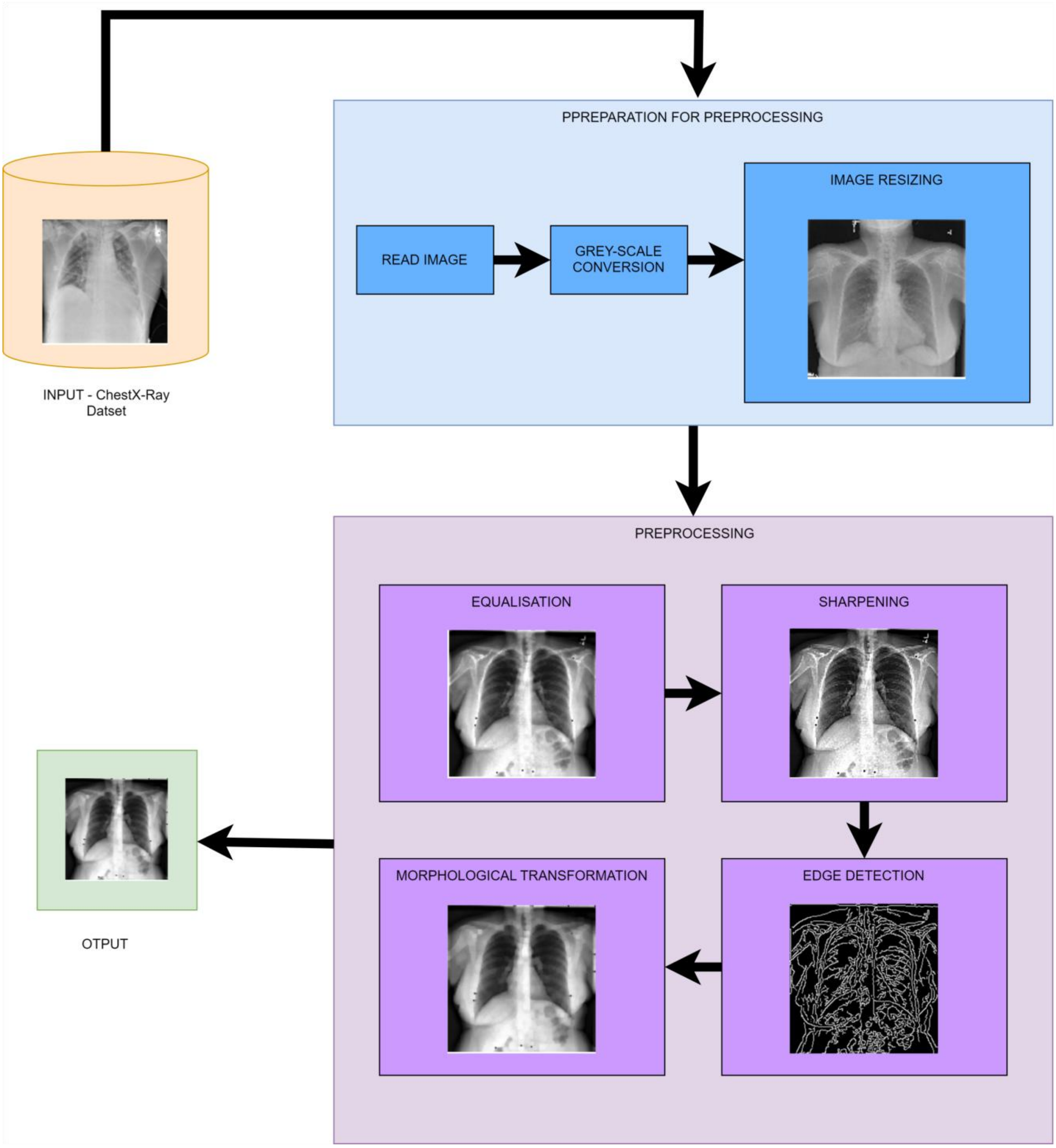

**Figure 4. Preprocessing Architecture**

The model architecture for preprocessing is depicted in Figure 4. This designs the preprocessing of the raw X-ray images in feature extraction throughout the process. The images are first digitized in grayscale so that they can be resized into square uniform sizes, then subject it to contrast enhancement through histogram equalization, sharpening for better clarity before morphological techniques such as dilation, erosion, edge detection using Canny and Sobel filters to enhance the anatomical features were applied. These techniques ensure that the model emphasizes the important features from the images and disregards noise.

This algorithm enhances images of chest x-rays of patients by applying several image processing techniques in series. The effect is achieved by first converting the image to gray scale by mixing the red, green, and blue channels. Then the images undergo re-sizing and the enhancement of its contrasts is done by histogram equalization. A filter is then applied for sharpening of detail, and subsequently the Canny method was used for edge extraction.

This morphological opening, involving first the erosion of the image then its dilation to eliminate noise and to enhance detail, is the focal area. Finally, the analysis of variance and edge energy will measure the effectiveness of the preprocessing steps. This all-encompassing approach is directed towards tending to many variables.

$$O_R^{(n)}(F) = R_F^D[(F \ominus nB)] \tag{1}$$

where $[(F \ominus nB)]$ represents a marker image and F is a mask image in morphological reconstruction by dilation and the (1) also represents an operation in mathematical morphology.$O_R^{(n)}(F)$ computes the opening of image $F$ with a structuring element $nB$ under the structuring element $R_F^D$, where opening removes minor details while keeping bigger structures. $\ominus$ implies erosion and $\ominus$ denotes reconstruction.

$$R_F^D[(F \ominus nB)] = D_F^{(k)}[(F \ominus nB)] \tag{2}$$

(2) determines the image $F$ dilation with the structuring element $nB$ under that condition when $D_F^{(k)}$ is equal to $k$, which means $k$ iterations of dilation. Dilation expands the image lines that have their shape as a prerequisite for many image processing tasks such as edge detection and feature extraction and D denotes geodesic dilation with k iterations until stability such that,

$$D_F^{(k)}[(F \ominus nB)] = D_F^{(k-1)}[(F \ominus nB)] \tag{3}$$

$$A \cdot B = (A \oplus B) \ominus B \quad (4)$$

where (4) represents the formula for closing of a set A by a structuring element B is the erosion of the dilation set.

$$A \oplus B = \cup_{b \in B} A_b \quad (5)$$

(5) describes the morphological dilation operation, where $A \oplus B$ can be thought of as set $A$ with all translations of set $B$ placed on top of it. It is the basis of $A$ with multiple copies of $B$ translated for this purpose, the morphological image processing tasks which involve large picture structures and $A_b$ is the translation of A by b

$$A \oplus B = \{z \in E \mid (B^s)_z \cap A \neq \emptyset \quad (6)$$

In (6) $A \oplus B$ is the dilation morphism defined on $(E, A)$ with $B \subset E$. The union of elements in$E$ is calculated for which the structuring element $B^s$ is already crossing set $A$, making the effect of the morphology on the picture structures broader and more significant

$$B^s = \{-x \in B\} \quad (7)$$

(7) expresses the symmetry of element B in space E. $B^s$ is the set of elements $x$ of $E$whose opposite $(-x)$ belongs to $B$. It is ensured that $B^s$ remains symmetric, and symmetrical operation is very important for morphological processes in image processing, for instance, erosion and dilation.

The erosion of the image A by the structuring element B is defined by (8) that is,

$$A \ominus B = \{B_z \subseteq A\} \quad (8)$$

(8) is about morphological erosion, $A \ominus B$, where $A$ gets smaller by using $B$ with respect to the space $E$. The set includes elements in $E$ where the translated structural element $B_z$ is entirely belonging to set $A$, this set is useful in image processing tasks to reduce small objects while keeping large entities and $B_z$ is the translation of B by the vector Z that is,

$$B_z = \{b \in B\}, \forall z \in E \quad (9)$$

(9) stands for the apparent translation of the structure element $B$ in the space $E$. $B_z$ is the result of adding each member of $B$ to every element in $E$, thereby uniformly distributing$B$over the entire $E$.

$$A \ominus B = \cap_{b \in B} A_{-b} \quad (10)$$

(10) shows that morphological erosion is obtained by $A \ominus B$through which set $A$ is contracted by structuring element $B$. It performs translations of $A$ by the elements $b$ in $B$ negated, intersecting them and shrinking $A$ to a reduced version that retains only those areas where $B$ fully fits, which is crucial for feature extraction in image processing.

| **Algorithm 1: Morphological Opening for Image Enhancement** | | |
|---|---|---|
| | ***input:*** *thoracic radiographs of human subjects* | |
| | ***grey scale conversion:*** | |
| | | $gray_value \leftarrow \frac{0.299 \times r^2}{g+b} + log(0.587 \times g) + \sqrt{0.114 \times b} + \frac{r \times g \times b}{255^2}$ |
| | | ***colour channels****:* ***r*** *→red,* ***g*** *→green,* ***r*** *→blue* |
| | ***image resizing:*** | |
| | $I_{resized}(x', y') \leftarrow I_{original}\left(\frac{x'}{r_x}, \frac{y'}{r_y}\right)$ | |
| | $x', y'$ *→ coordinates of the resized image,* $r_x, r_y$ *→scaling factors of x and y* | |
| | ***histogram equalisation:*** | |
| | | $T(i) \leftarrow \frac{\sum_{j \leftarrow 0}^{i} n_j}{M \times N} \times L$ |
| | | $n_j$ *→histogram of input image,* ***M, N*** *→ dimensions of input image,*<br>***L*** *→ number of intensity levels* |

| | | | |
|---|---|---|---|
| | ***sharpening filter:*** | | |
| | $sharpened \ = I_{equalized} + k \times Laplacian(I_{equalized})$ | | |
| | $V \leftarrow \frac{\sum_{i \leftarrow 1}^{N} log(I_{equalized}(i))}{\sqrt{N}}$ | | |
| | $sharpened \leftarrow sharpened + V^2$ | | |
| | $I_{sharpened}(x, y) = \sum_{i=-1}{}^{1} \sum_{j=-1}{}^{1} I(x + i, y + j) \times K(i, j)$ | | |
| | *x, y → coordinate of pixels, i, j → coordinates within kernel matrix* | | |
| | ***edge detection:*** | | |
| | | $E_1 \leftarrow \frac{\sum_{i \leftarrow 1}^{N} Sobel(I_{equalized}, axis \leftarrow x)^2}{\sqrt{N}}$ | |
| | | $E_2 \leftarrow \left(\frac{\sum_{i \leftarrow 1}^{N} Sobel(I_{equalized}, axis \leftarrow y)}{N}\right)^3$ | |
| | | $edges \leftarrow Canny(I_{equalized}, threshold1 \leftarrow 30, threshold2 \leftarrow 100) + E_1 - E_2$ | |
| | *for i ← 0 to n do:* | | |
| | | ***morphological opening:*** | |
| | | | $morph = (I_{equalized} \ominus B) \oplus B$ |
| | | | $\ominus$→*erosion,* $\oplus$→*dilation* |
| | | | $M_1 \leftarrow \frac{\sum_{i \leftarrow 1}^{N} Erode(I_{equalized}, B)^2}{\sqrt{N}}$ |
| | | | $M_2 \leftarrow \left(\frac{\sum_{i \leftarrow 1}^{N} Dilate(I_{equalized}, B)}{N}\right)^3$ |
| | | | ***N*** *→number of pixels in the image* |
| | | | ***return*** *image* |

Thoracic radiographs of the human subjects are aimed to make a better outcome with this preprocessing approach. It includes a range of adjustment in order to increase the sharpness and the quality of the image.The first step is to get the input images into a grey scale. The grey scale is useful because it makes possible easy processing of the image by showing it in a single intensity channel. As a next step, the same conversion method would ensure the image data is consistently handled.Through scaling the photos using scaling factors they are changed to greyscale after conversion to greyscale.

This way the guarantee of uniform image sizes for different machines and datasets is obtained for appropriate processing on all systems. These images are then brightened and contrasted other ways, like histogram equalisation. By using the same technique, the method sorts the pixel intensities on the whole dynamic range in such a manner that the details in different intensities are enhanced.

Now, the edge-enhancing filter is applied to the image to emphasise the edges and elements of the image. The filter is set to draw attention to the changes between neighbouring pixel values, which in turn results in improved clarity of the image and the structures is better defined and crispier. After that, edge detection is used to point out edges and differentiate contours among the pictures. As a first important step, the algorithm uses this stage to distinguish anatomical landmarks and anomalies from each other.

The next stage is going through the morphological opening that helps to improve the quality of the photos. Erosion filter and dilation operations are combined in this operation to keep small scale noise while keeping the whole shape of features. Each step in the process as well as the mathematical and computational activities are designed to help improve the output images. The method is aimed at making the examinations of thoracic radiographs more comprehensible and interpretable through the successive implementation of multiple techniques that will eventually lead to a more precise diagnosis and analysis in medical imaging applications.

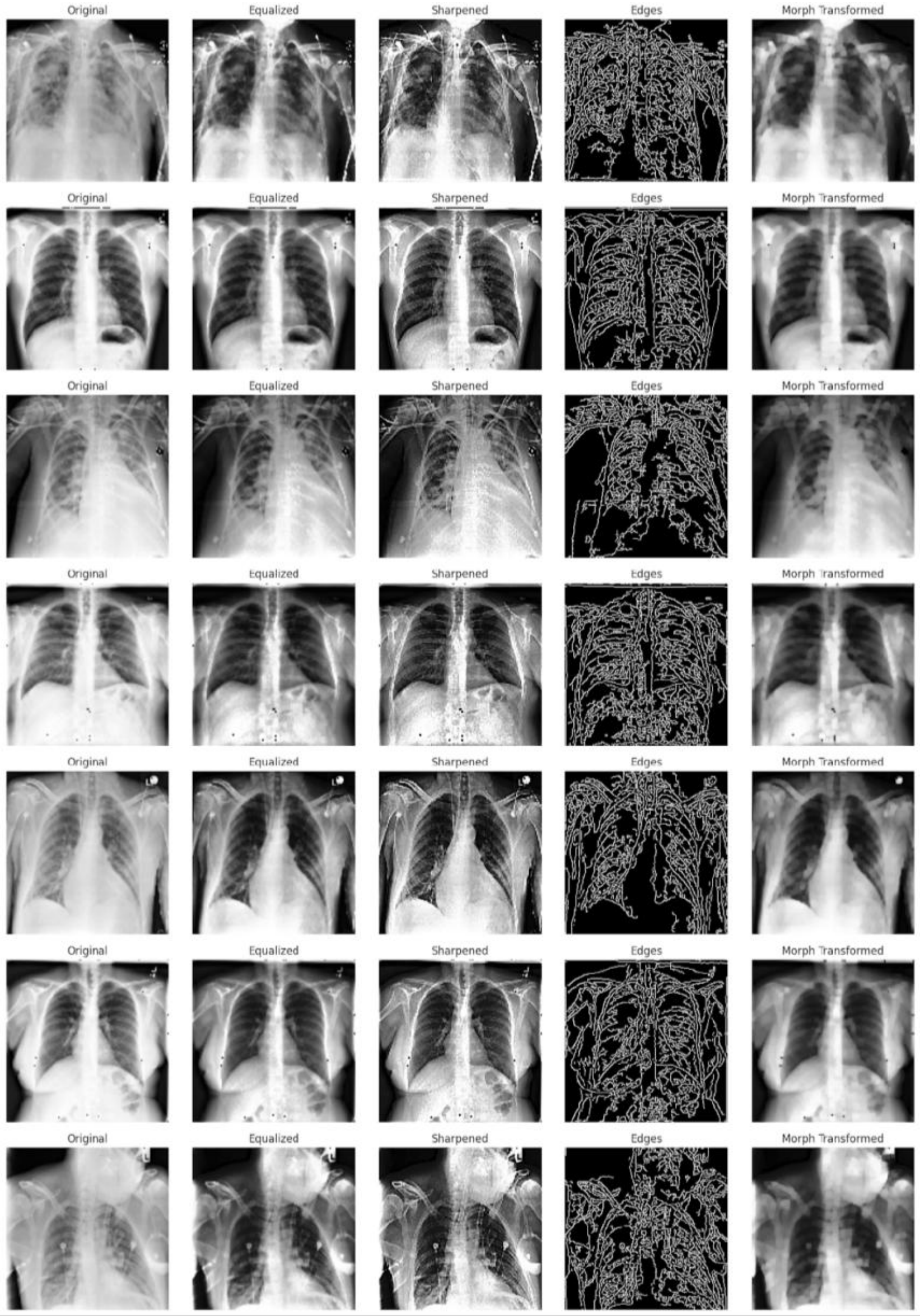

**Figure 5. Preprocessing Output**

Figure 5 indicates how different image processing techniques are employed to enhance X-ray scans for higher yields. Each row indicates an X-ray that has been processed through a series of processing operations, including edge detection sharpening for enhanced visibility of edges, morphological processing for enhanced structural detail, and contrast enhancing by histogram equalisation for enhanced contrast. Edge detection also identifies prominent anatomical structures, allowing the model to focus more on important patterns. Together, these preprocessing steps improve the model to be capable of identifying cardiomegaly accurately.

## 3.3 Classification

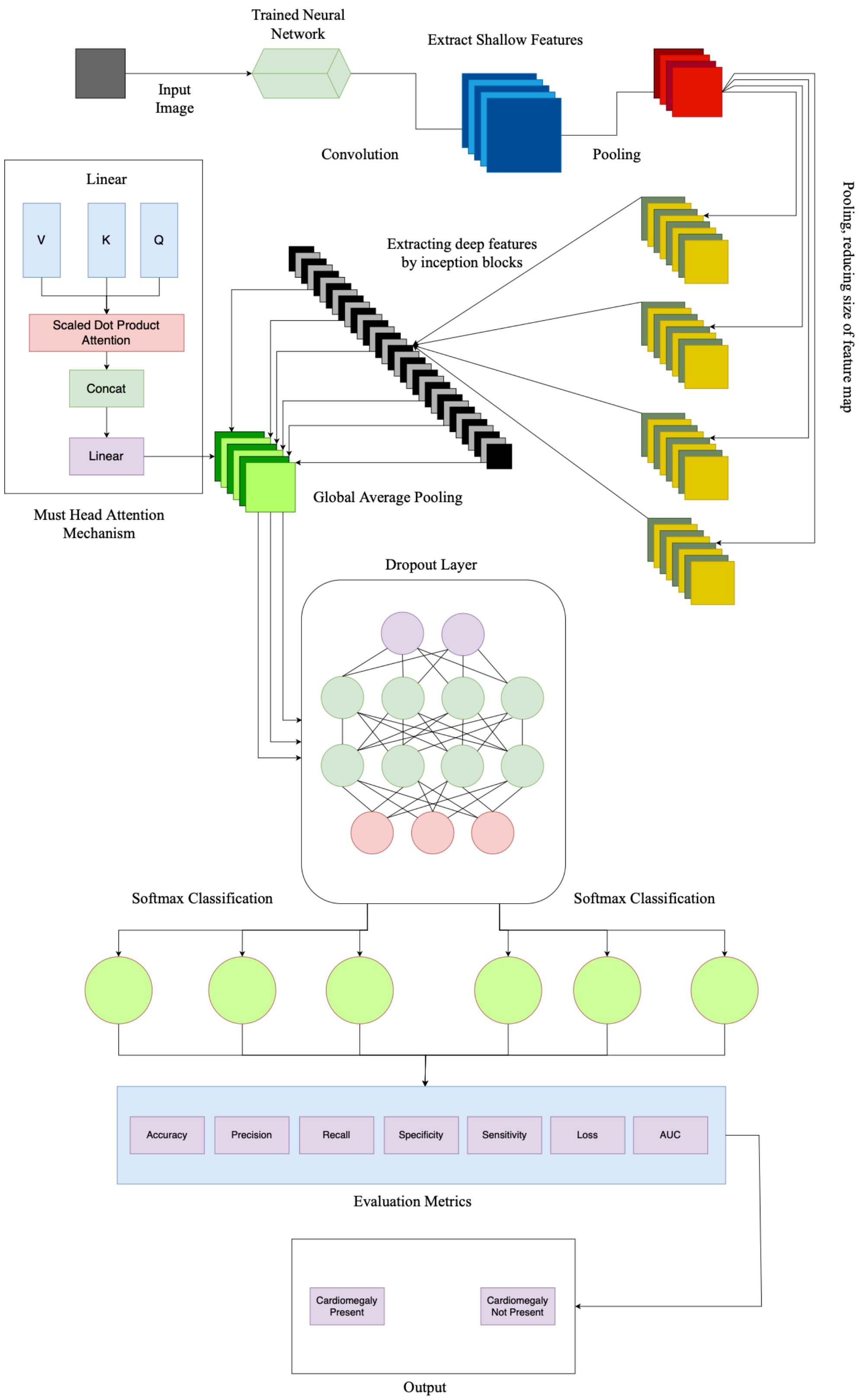

**Figure 6. Overall Classification System Architecture**

Figure 6 tells us about the architecture of the classification system with both CNN-based feature extraction and attention for enhanced performance. The model starts with preliminary visual pattern extraction with convolutional and pooling layers followed by feature representation improvement with inception blocks. Data reduction is performed with salient details maintained with a global average pooling layer. Multi-head attention mechanism improves the model's attention to the correct places in the X-ray to give improved interpretability. Dropout layers are added for preventing overfitting, and the final classification layer is given a softmax activation function to predict the probability of cardiomegaly occurrence. The efficiency of the model is validated with the principal measures such as accuracy, precision, recall, specificity, sensitivity, loss, and AUC.

With its several variants based on several parameters, the classification module is the final decision-making module in the proposed deep learning architecture, combining the feature representations acquired through InceptionV3 and refined through the multi-head attention system. This module renders high-dimensional feature representation down to meaningful predictions of classes, declaring whether or not cardiomegaly is present in a chest X-ray. The stages contributing to the classification process are dimming, regularization, feature transformation, and estimating the estimated probabilities. Each of these processes is implemented specifically in a way that would enhance the model performance by keeping significant pieces of information intact and ready to reduce overfitting and bolstering the robustness of predictions.

The proposed method uses Global Average Pooling (GAP) that reduces the spatial feature maps extracted by means of the convolutional layers and attention mechanism into lower-dimensional feature representations. Unlike traditional fully connected layers that treat the flattened feature maps with their general management, the pools are able to avoid the destruction of spatial information by averaging the activation of each feature channel. This model has definitely less computation compared to those with FCs, and it has also been noted that the forces it induces act in a manner reminiscent to regularizing. By that, GAP ensures most relevant information is kept by searching covariates through space aggregation by combining feature responses in spatial dimension while simple redundant activations are dropped, leading to a much more variance-reduced representation of learned features.

After the GAP layer, the model has a dropout layer that regularizes the model for improved generalization. Dropout is a procedure that, during training, is randomly switched off for a few neurons, producing noise in the network such that it is discouraged from being overly reliant on particular feature activations. This stochastic technique enforces the model to learn many diverse and distributed feature representations, and in this way reduces its susceptibility to overfitting. The probability of deactivating the neurons is tuned so as to give a suitable trade-off between regularization and model capacity, ensuring that the network is expressive enough while robust against unseen data. During inference, dropout is discarded, and all neurons contribute to the final prediction, using the entire joint power of the information represented.

Output of the dropout layer is passed through a fully connected (dense) layer, which thus serves as the final transformation stage before classification. This dense layer integrates the processed features into a high-level decisioning representation mapping the extracted features to a latent space where class separability is maximized. The weights of this layer are learned through backpropagation, allowing the model to fine-tune its decision boundaries in light of the training data. The dense layer mostly serves to bring the fusion of features, infusing multiple attention-enhanced feature channel information into a unified construct of diagnosis that reflects both deep and local artistic patterns indicative of cardiomegaly.

The outputs generated by the model are finally scored using a softmax activation function. For a model that attempts to classify the presence or absence of cardiomegaly, the output will be two classes, categorical in nature. The softmax function transforms the output from the dense layer to a set of probabilities that sum to one, enabling the model to make a judgment regarding classification according to the highest probability score. The softmax function calculates probabilities by invoking the exponential over all logits (raw output from the dense layer) in order to normalize them. Each output now represents a confidence score (from 0% to 100%) for each class. Such a probabilistic interpretation becomes extremely useful in the clinical diagnostics setting, where the operational thresholds for decision-making could vary based on clinical requirements, for instance, under conditions where sensitivity must be prioritized over specificity in the treatment of false-negative cases.

In its aim to optimize the classification module, the model explicitly incorporates the binary cross-entropy loss function, which is best suited for binary classification tasks. Binary cross-entropy describes the perceived distinction between predicted and actual class probabilities;

giving a greater penalty for an incorrect class prediction, particularly where the prediction is made with a high degree of confidence. The loss function thus encourages the model to remain as deterministic as possible about its predictions, progressively improving the classification accuracy over the various training iterations. To optimize the model's performance, the RMSprop algorithm was used, again considering the adaptive learning rate. RMSprop prevents huge weight changes due to its working being dependent on the mean Moverage of squared gradients, so it stabilizes against any abrupt changes during convergence, even while the learning dynamics are highly non-stationary.

Various metric scores will be obtained to evaluate the performance of the classification module: accuracy, precision, recall, specificity, sensitivity, loss, and AUC. Accuracy refers to the number of correct predictions made by the model overall. Precision and recall together show the technical competence of a model in identifying all positive instances. Model-specific sensitivity refers to the ability of a model to effectively separate positive cases from negative, thus minimizing the chances of false positives. Sensitivity is considered one aspect of highest importance, especially in this case, due to the consideration of its identification in actual cases of cardiomegaly dependent on its clinical application scope. AUC is a measure of model performance summarized across different levels of sensitivity and specificity based on the trade-offs models are making at each point of the chosen decision threshold.

The proposed model, by offering strength of discrimination, robustness, and clinical validity through enhancement of existing classification methods, achieves one clearer cut-off from the other in its predictions. In other words, attention-based feature representations, dropout-based regularization, and softmax-based decision generally contribute to a more accurate classification with fewer false positives. A specific degree of model capture for minor differences in the presentation of cardiomegaly guarantees itself to be quite a dependable diagnosis and really useful for it.

$$MultiheadedAttention(Q,K,V) = Concat_{i\in[\#heads]}\left(Attention\left(XW_i^Q, XW_i^K, XW_i^V\right)\right)W^O \quad (11)$$

where X is the concatenation of word embeddings, and the matrices $W_i^Q, W_i^K, W_i^V$ are projection matrices owned by individual head $i$ and $W^O$ is a final projection matrix owned by the whole cult head attention head

$$MaskedAttention(Q, K, V) = softmax\left(M + \frac{QK^T}{\sqrt{d_k}}\right)V \quad (12)$$

(12) is Masked Attention operation, which is the mechanism of neural networks. It computes attention scores by scaling dot product of query $Q$ and key $K$ which is added to a masking matrix $M$ and a softmax is applied. The resulting attention weights are applied to weigh for the $V$ values, giving the resultant context-based information extraction.

$$w_{k+1} = w_k - \alpha \nabla f(w_k) \quad (13)$$

(13) represents the Stochastic Gradient Descent (SGD). The the weights are nudged in the negative gradient direction. Despite its simplicity, good results can still be obtained on some models.

$$z_{k+1} = \beta z_k + \nabla f(w_k) \quad (14)$$

(14) shows that it is possible to optimize the movement pattern depending on the momentum. The expression $z_{k+1}$stands for the updated term of momentum, which multiplies the momentum coefficient $\beta$, and the gradient of the objective function at the moment of parameter weight $w_k$ This method applies both the past momentum as well as the present slope, thereby enhancing the rate of convergence.

$$w_{k+1} = w_k - \alpha z_{k+1} \quad (15)$$

The formula describes a recursive update rule in optimisation algorithms, where $w_{k+1}$ represents the updated parameter vector, $\alpha$ signifies the step size, and $z_{k+1}$ represents the gradient of the objective function. The objective is to modify the parameter vector in order to minimise the objective function. (14) and (15) describes the updates performed by momentum optimizer, which can also be written as follows

$$w_{k+1} = w_k - \alpha \nabla f(w_k) + \beta(w_k - w_{k-1}) \quad (16)$$

the last term in (16) is the component in the direction of the previous update and portrays a means of optimizing a value which is gotten from combining the methods of gradient descent and momentum. A new parameter vector $w_{k+1}$ is created as the gradient $\nabla f(w_k)$ is scaled and the momentum from the previous step $w_{k-1}$ is added; the latter is scaled by$\beta$ and the step size $\alpha$ is also added. This way, the algorithm achieves a higher speed to the solution and stability of the solution is ensured.

$$g_{k+1}^{-2} = \alpha g_k^{-2} + (1-\alpha) g_k^2 \tag{17}$$

(17) is used to figure out the exponentially weighted moving average of the squared gradients in optimisation algorithms. The squared gradient inverse, $g_{k+1}^{-2}$, takes a weighted sum of the current squared gradient, $g_k^2$, and the previous estimate, giving: $g_{k+1}^{-2} = (1-\rho) g_k^{-2} + \rho g_k^2$. The weight is decided by the measurable \( \alpha \) factor. This contributes to step sizes change adaptation capability during the optimisation process.

$$w_{k+1} = \beta w_k + \frac{\eta}{\sqrt{g_{k+1}^{-2}{}^{+\epsilon}}} \nabla f(w_k) \tag{18}$$

The formula makes the vector $w_k$ of parameters change for optimisation methods. The integration of a momentum term $\beta w_k$ and the dependence of the step size on $\nabla f(w_k)$divided by the square root of the moving average of squared gradients $g_{k+1}^{-2}{}^{+\epsilon}$. This way, parameter adjustment as well as convergence improvement are achieved. (17) and (18) also describes the working of RMSprop which is the optimizer used in the classification algorithm

| **Algorithm 2: Multi-Head Channel Attention Integrated with Inception V3 for Cardiomegaly Classification** | | |
|---|---|---|
| | ***input:*** *thoracic radiographs obtained from preprocessing* | |
| | ***base_model*** *→ Inception_V3 with weights extracted from* ***ImageNet*** | |
| | ***for*** *i ← 0 to tot_layers* ***do*** | |
| | | ***froze_layers*** *→ set trainable to* ***False*** |
| | ***function(****multi_head_attention, args→****input, heads):*** | |

| | | | |
|---|---|---|---|
| | | $$model_depth \leftarrow \frac{input_shape[-1]}{heads}$$ | |
| | | ***for*** *i ← 0 to heads* ***do*** | |
| | | | $$attention_scores \leftarrow Q \cdot K^T$$ |
| | | | $$MultiHeadAttention(Q, K, V) = Concat(head_1, head_2, \ldots, head_h) \cdot W^O$$ |
| | | | $$attention(Q, K, V) \leftarrow softmax\left(\frac{QK^T}{\sqrt{d_k}}\right)V$$ |
| | | | $$scaled\ attention\ scores \leftarrow softmax\left(\frac{attention_scores}{\sqrt{depth}}\right)$$ |
| | | | $$head_i \leftarrow log(abs(QW(KWK)_iT + \epsilon)).(VW)^T$$ |
| | | | $$concatenated_output \leftarrow Concatenate(outputs)$$ |
| | | | ***return*** *attention output* |
| | ***function(****create_model, args→****Null):*** | | |
| | | $$input_layer \leftarrow shape(x, y, z)$$ | |
| | | $$convolution_output(i, j) \leftarrow \sum_m \sum_n input(i + m, j + n) \times kernel(m, n)$$ | |
| | | $$attention_output \leftarrow multi_head_attention(base_model)$$ | |
| | | *pass* ***attention_output*** *through global average pooling (2d)* | |
| | | $$global_average_pooling \leftarrow \frac{\sum_i \quad x_i}{num_channels}$$ | |
| | | $$For\ each\ channelc: GAP_c \leftarrow \frac{1}{H \times W} \sum_{i \leftarrow 1}{}^{H} \sum_{j \leftarrow 1}{}^{W} X_{ijc}$$ | |
| | | ***H*** *→ height of the feature map,* ***W*** *→ width of the feature map*<br>***C*** *→ number of channels* | |
| | | ***dense layer activation:*** | |
| | | *f(x) ← max(0,x)* | |
| | | ***loss function:*** | |
| | | $$L(y, \hat{y}) \leftarrow -\sum_i y_i log(\hat{y}_i)$$ | |
| | | $L(y, \hat{y})$ *→ denotes the loss between the true labels and the predicted probabilities*<br>$y_i$ *→ true label of class i*<br>$\hat{y}_i$ *→ predicted probability of class i* | |

| | | |
|---|---|---|
| | | ***optimizer:*** |
| | | $$w_{t+1} \leftarrow w_t - lr \times \frac{m_t}{\sqrt{v_t} + \epsilon}$$ |
| | | $w_{t+1}$ → *updated parameter at time t + 1,* $m_t$ → *first moment estimate of gradients* $v_t$ → *second moment estimate of the gradients* |
| | | ***return*** *model* |
| | ***compute*** *accuracy* | |
| | | $$accuracy = \frac{\sum_{k\leftarrow 1}^{n} TP_k + TN_k}{\sum_{k\leftarrow 1}^{n}(TP_k + TN_k + FP_k + FN_k)}$$ |
| | ***compute*** *precision* | |
| | | $$precision = \frac{1}{n}\sum_{k\leftarrow 1}{}^{n} \frac{TP_k}{TP_k + FP_k}$$ |
| | ***compute*** *recall* | |
| | | $$recall = \frac{\sum_{k\leftarrow 1}^{n} \quad TP_k}{\sum_{k\leftarrow 1}^{n}(TP_k + FN_k)}$$ |
| | ***TP*** → *True Positive,* ***TN*** → *True Negative,* ***FP*** → *False Positive,* ***FN*** → *False Negative* | |
| | ***compute*** *Dice Coefficient* | |
| | | $$dice\ coefficient = \frac{2 \times \lvert A \cap B \rvert}{\lvert A \rvert + \lvert B \rvert}$$ |

The system utilizes thoracic radiographs to classify cardiac enlargement, integrating the Multi-Head Channel Attention mechanism with pre-trained Inception V3 architecture. Thoracic radiographs are the first data upon which this algorithm acts, with preprocessing already having been performed prior to being submitted for computation. Inception V3 is a convolutional neural network that has already been pre-trained upon the ImageNet dataset and is notably precise in image classification problems, making it the central model in our research.

The Inception V3 model's layers are stacked, and the algorithm trains them by freezing to halt the process. Doing this will help in not forgetting the traits which were learned and they will definitely be used in the subsequent training cycle. In addition, the technique creates a Multi-Head Channel Attention function that enables the channel attention mechanism to selectively perceive various feature map channels to enhance feature representation. The attention mechanism enables the program to extract more richer features, which is achievable through the iterative process involving multiple heads where each pays attention to a certain aspect of the input data.

The second function is to build a model that is generated by applying the attention output with a 2D global average pooling. The aggregation of the feature maps across the spatial dimensions on its own captures a dense representation that is adequate for further classification. To complete the model building, we employ a dense layer with ReLU activation. The configuration of the optimiser and the computation of the loss function are set next. This optimizer depends on the first and second order estimations of the gradients to update the model parameters based on the gradient descent principle, while the loss function suggests the difference between the actual labels and the expected probabilities. The accuracy, precision, recall, and dice coefficient are calculated once the model is generated, and system calculates the metrics. These measures are the evaluation methods for the model through demonstrating its precision in cardiomegaly case classification. As far as precision and recall, the capacity of the model to minimize false positives and false negatives is highlighted, whereas the measure of accuracy is the overall correctness of predictions. The largest value of the Dice coefficient's accuracy measure here is the spatial overlap between the expected segmentation and the actual segmentation.

## 4. Results and Discussion

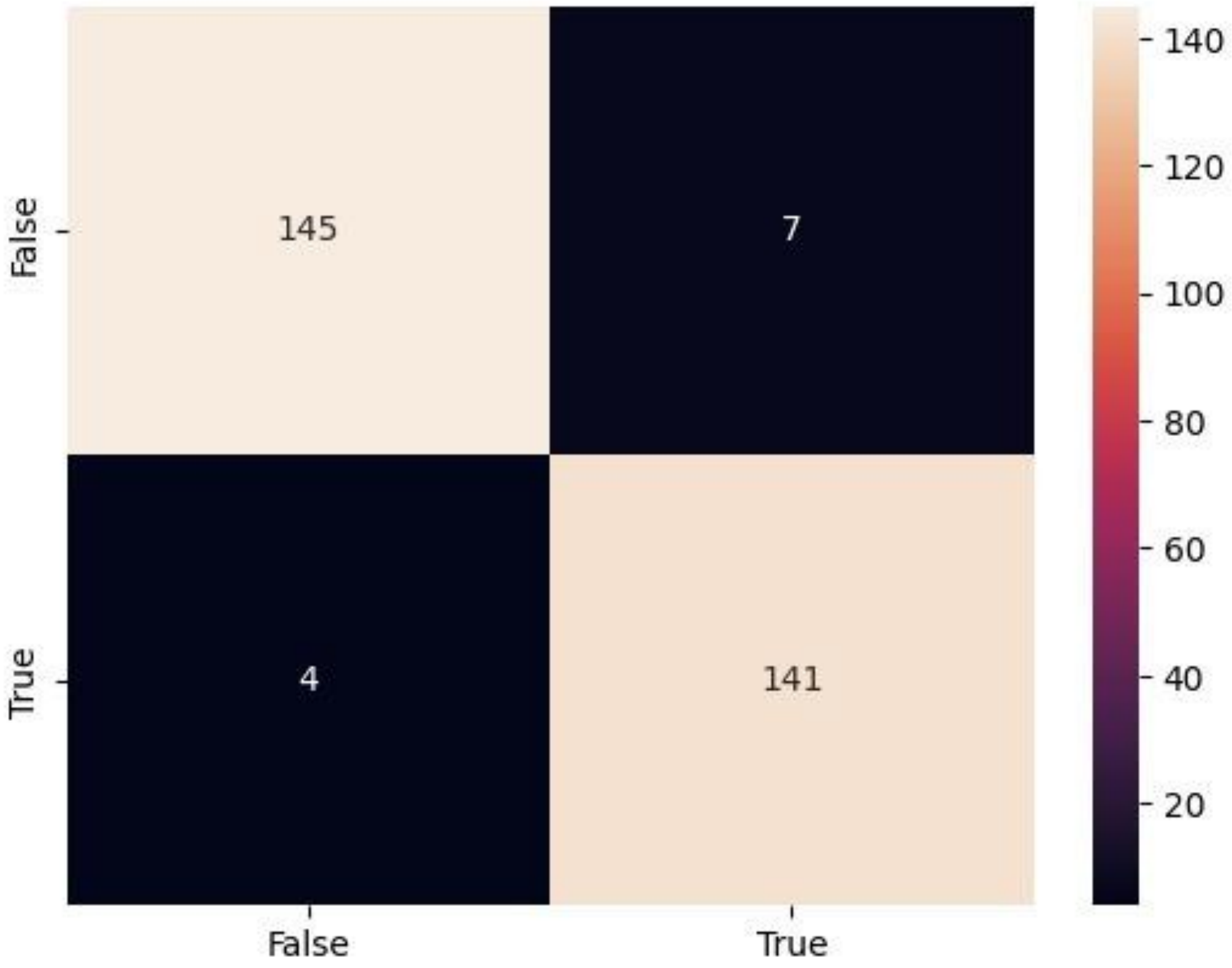

**Figure 7. Confusion Matrix**

Figure 7 presents the confusion matrix and model classification performance of the model. The model outputs show that there are 141 true positives, 7 false positives, 4 false negatives, and 145 true negatives with high accuracy. The minimal misclassifications reflect a high reliability, which is extremely important in diagnostic medicine. Here, false negative minimization is important not to overlook cardiomegaly cases, thus the confusion matrix being an important element to determine the model's reliability.

**Table 2. Confusion Matrix Summary**

| **True Positive** | 141 |
|---|---|
| **True Negative** | 145 |
| **False Positive** | 7 |
| **False Negative** | 4 |

Table 2 summarizes the confusion matrix, presenting the performance of this classification model. The proposed method correctly classified a total of 141 as true positive and 145 as true negative, with seven false positives and four false negatives; strong evidence of a reliable diagnosis lies within the confirmation of cardiomegaly, combined with an effective result.

Our proposed algorithm’s incredible performance is certainly a testament to its future application in real-life settings, for example in medical diagnosis where the ultimate goal is a precise classification. By the analysis of the confusion matrix we approach the problem of the model’s performance in classifying and discriminating between various classes more deeply.

**True Positives (TP)**: The model, which picked up 141 positive samples from the dataset, has well-classified the target class. These are situations where the model is exactly right in the sense that it predicted the presence of the target condition the way it should be, showing that it is capable of identifying relevant patterns and making correct classifications.

**True Negatives (TN)**: There were 145 cases of the negative category of the target condition that were correctly classified, and thus, the model could distinguish absence of the target condition well. This points to the model's capability to distinguish between instances that are not from the positive class and hence indirectly contribute to the model's classification precision

**False Positives (FP)**: There were seven disordered cases that have been falsely classified as positive but they are actually in the negative class. These false alarm cases are cases in which the model wrongly warned about the target condition and as a result, the real situation is misinterpreted or it led to wrongful interventions. False positives must be cut down because in such application where accuracy is essential, inappropriate consequences can not be allowed.

**False Negatives (FN)**: The model misclassified as negative the four cases that are true positives. Such cases stand for uncaught blunders, which could imply the errors in classification, poor analysis of important data or not providing the needed interventions. The most crucial thing about reducing false negatives is that its significance cannot be overstated,

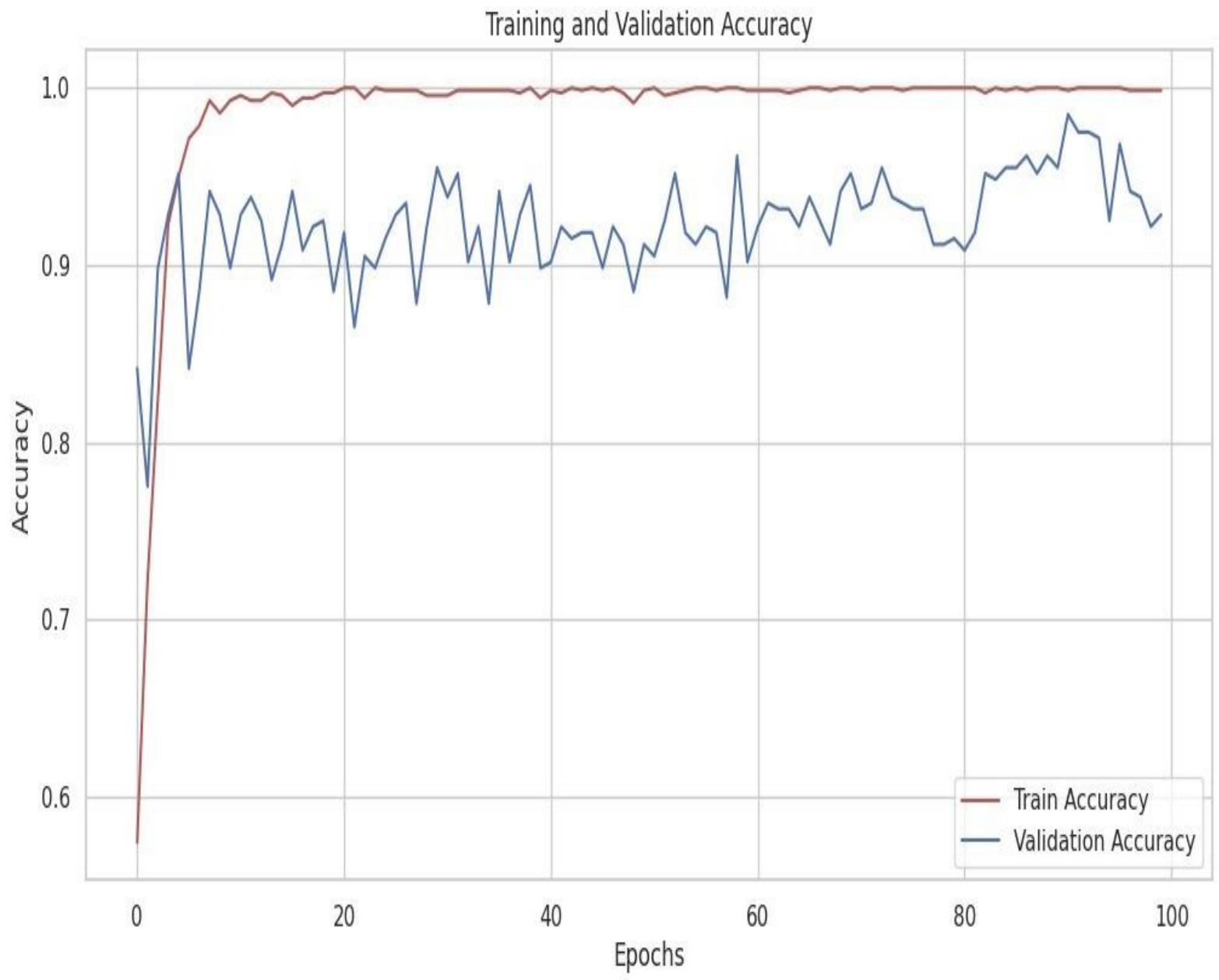

**Figure 8. Accuracy Curve**

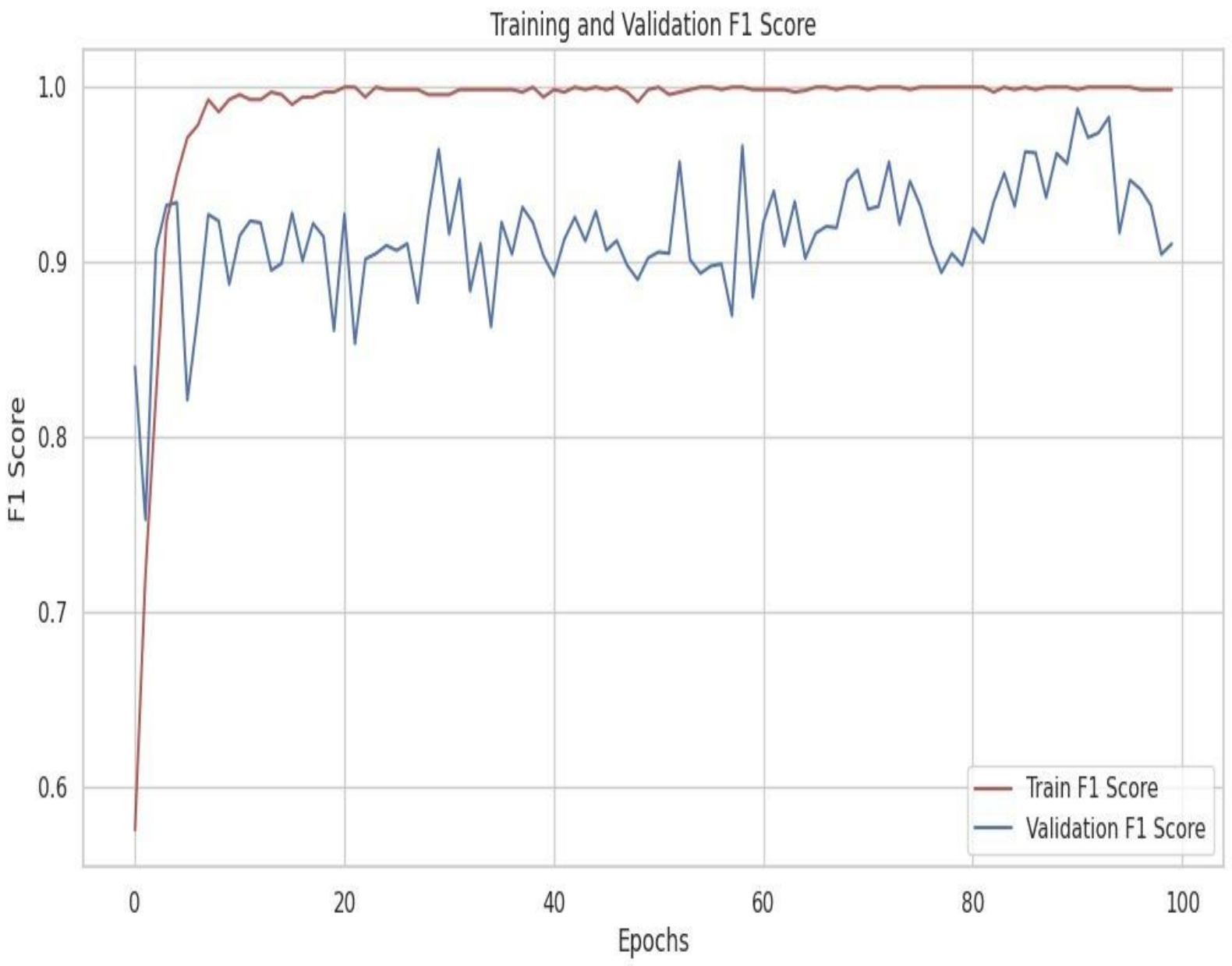

**Figure 9. F1 Score Curve**

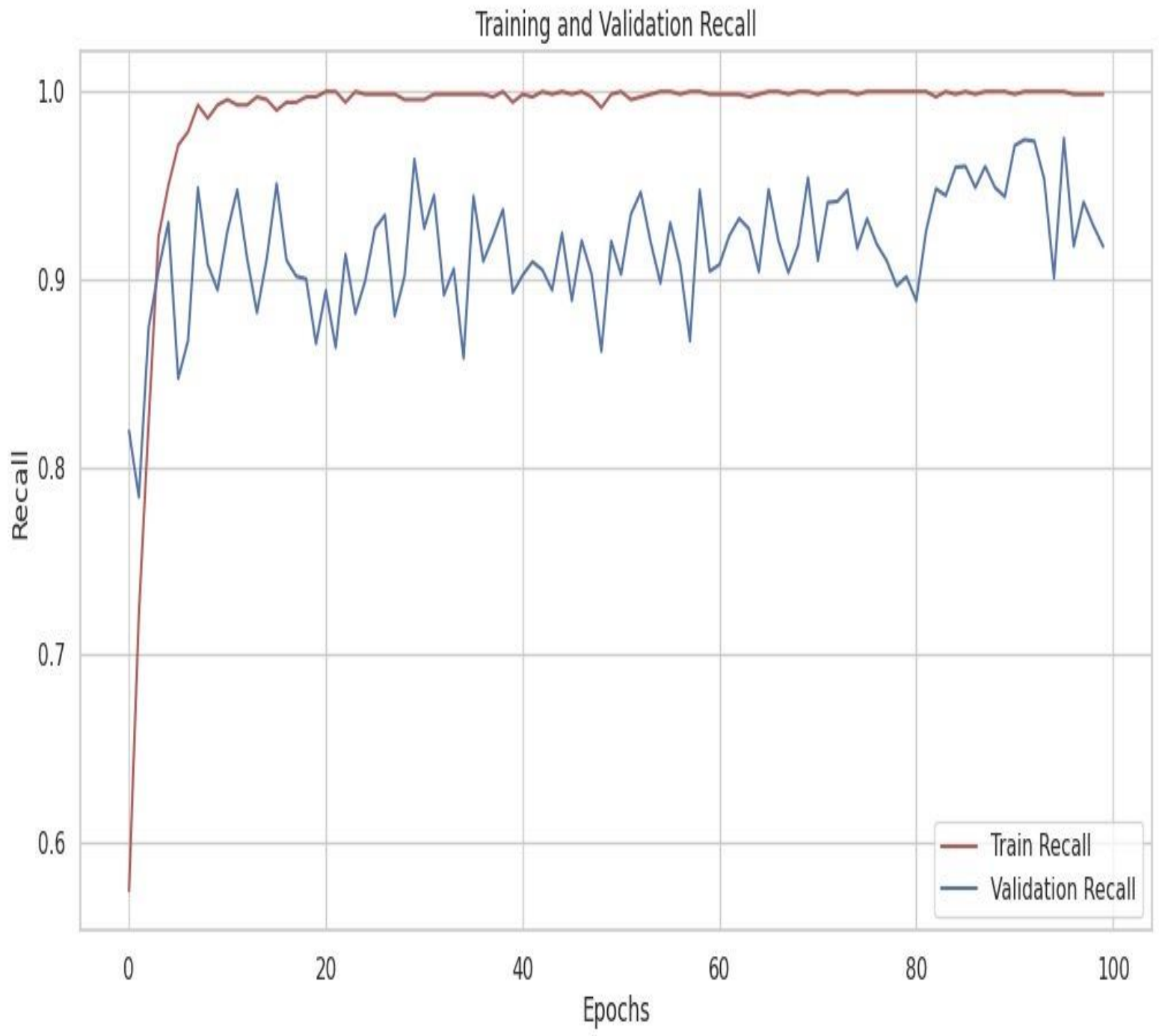

**Figure 10. Recall Curve**

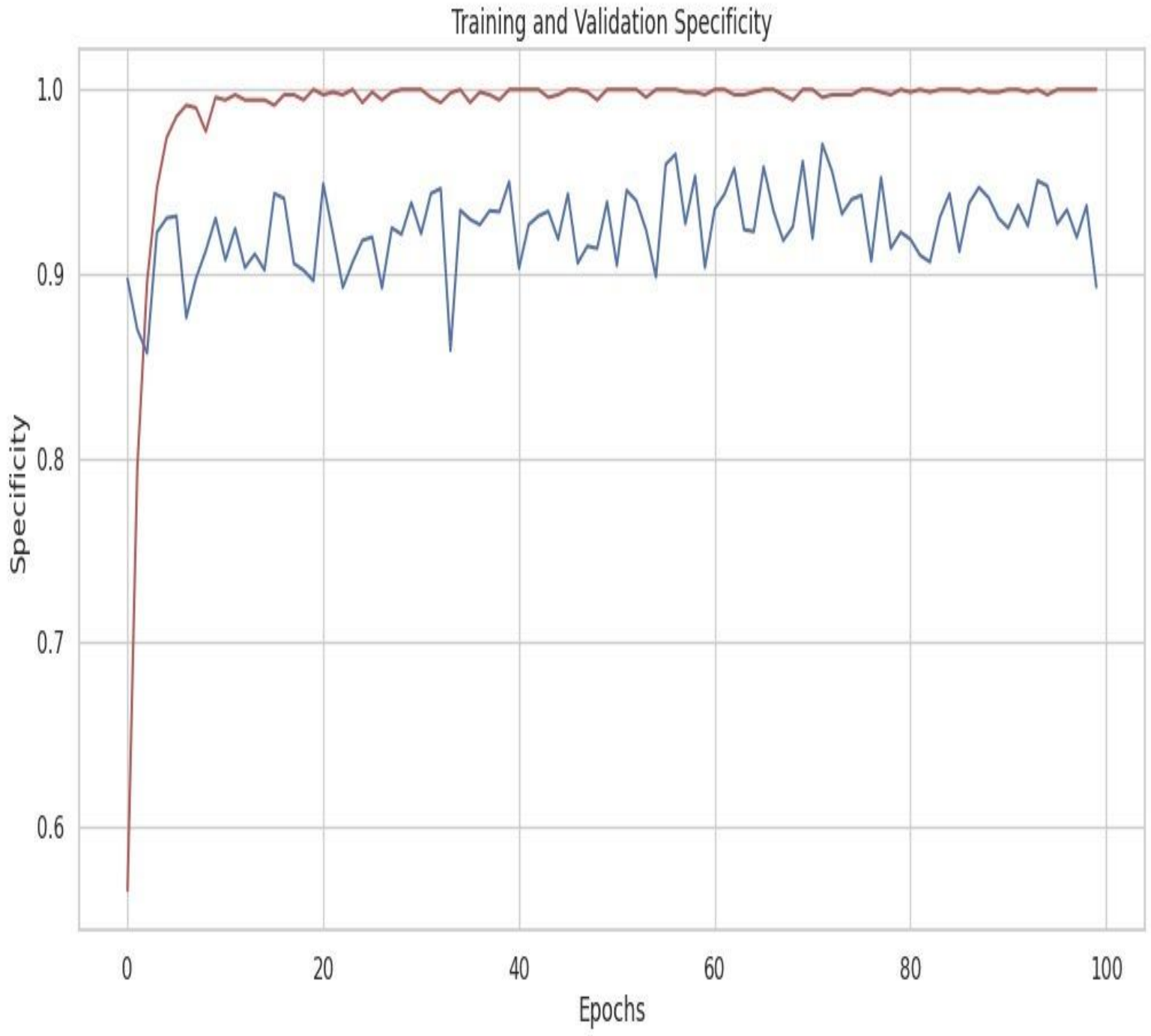

**Figure 11. Specificity Curve**

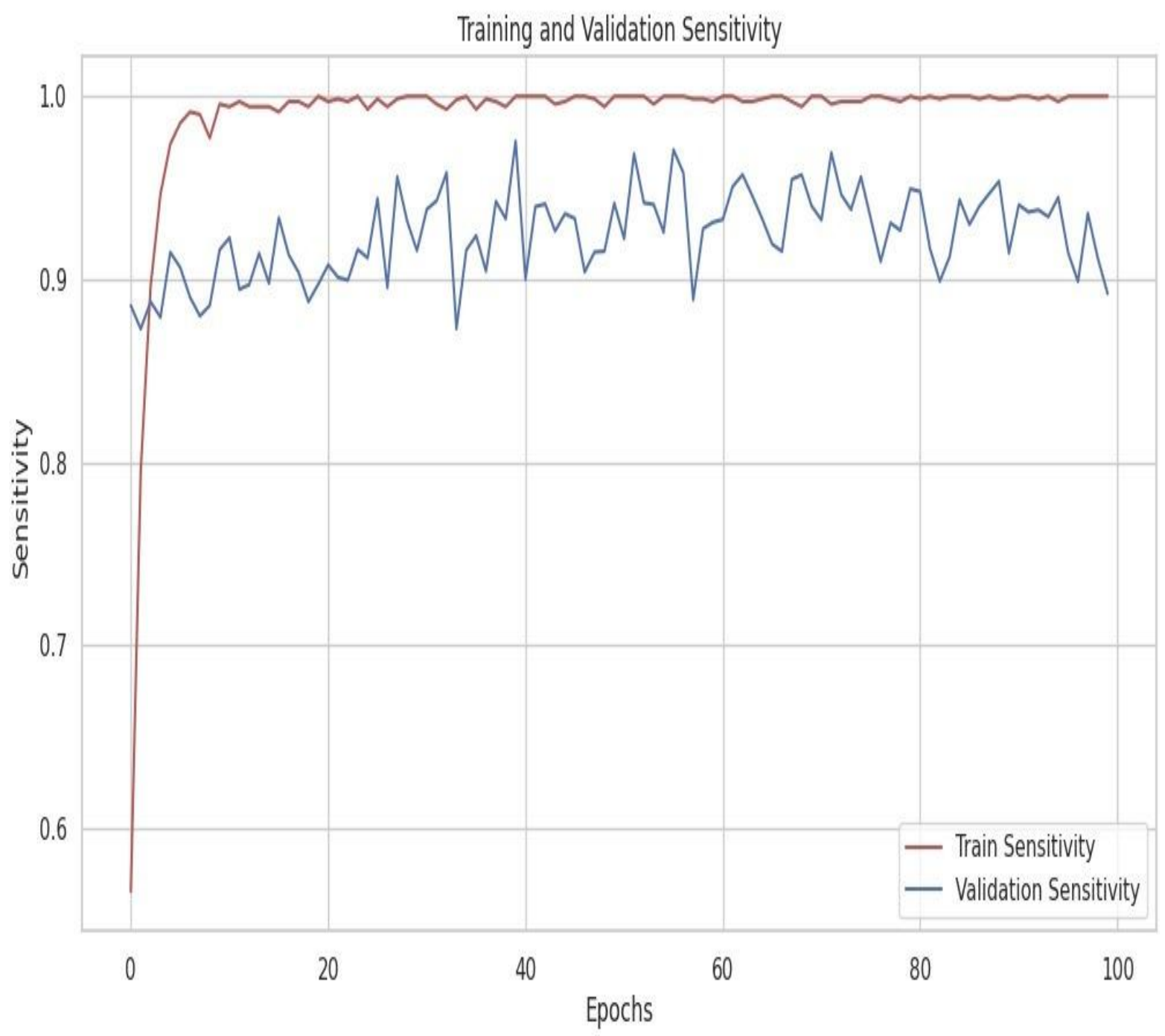

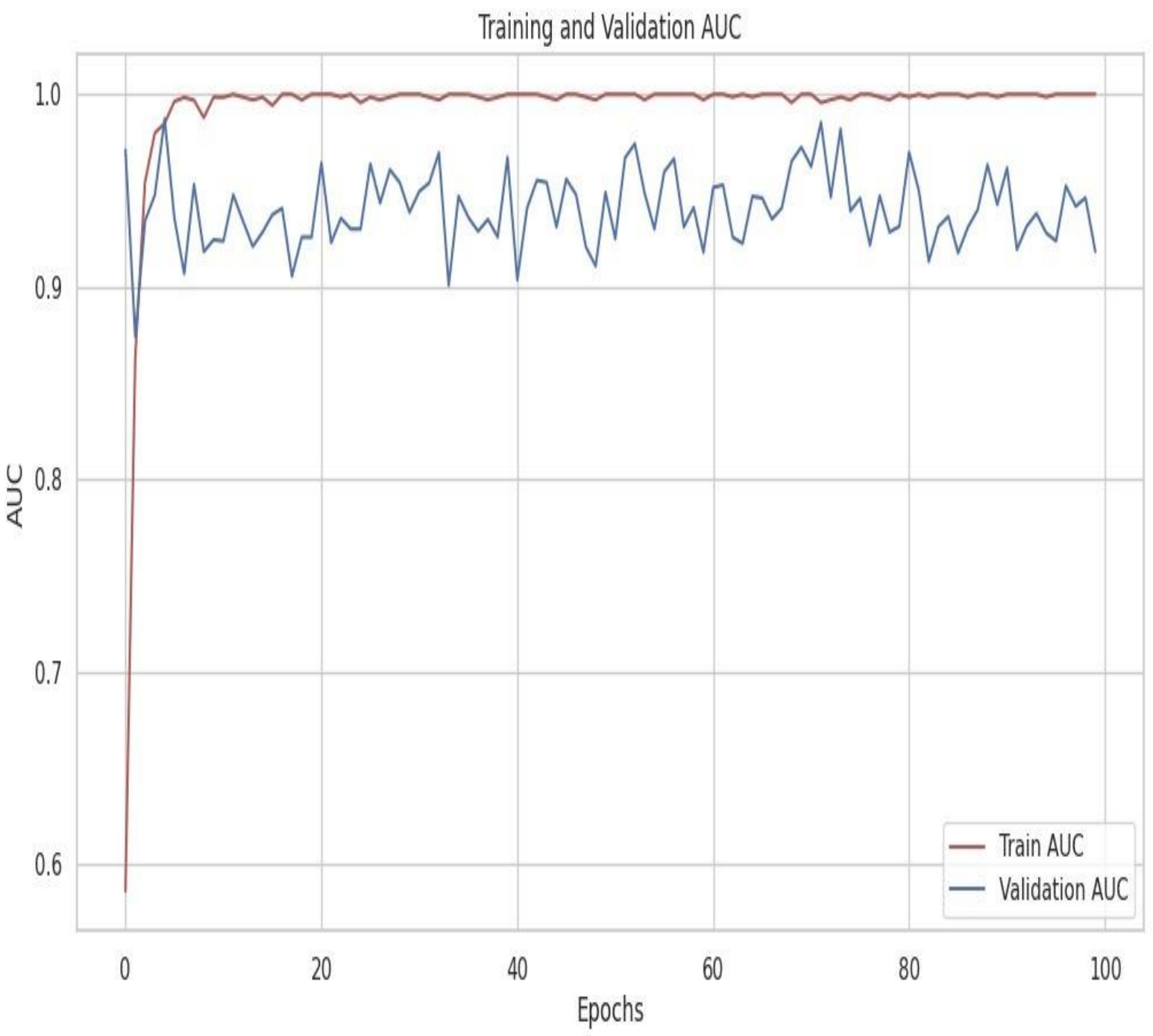

**Figure 12. Sensitivity Curve** **Figure 13. AUC Curve**

Figure 8 shows the training versus validation accuracy trends over 100 epochs for the proposed model. The rapid convergence was achieved with validation accuracy close to 95.6%. Minimal divergence in training and validation accuracy indicates relatively low overfitting and a high generalization ability, thus applicable for cardiomegaly detection in the real-world.

Figure 9 presents the trends in training and validation scores respectively of F1—an indicator measuring the balance between precision and recall. Above 90%, the validation F1-score becomes stabilized and accentuates the model's capacity to classify positive cases correctly while minimizing false-positive and false-negative errors. Such performance backed by medical diagnosis reaffirms the reliability of the model.

Figure 10 depicts recall performance by the proposed model throughout the training epochs. Recall gives a ratio of true positives correctly identified within the set of actual positive cases. The validation recall being more than 90% indicates that the model is effective in detecting cardiomegaly cases while minimizing false negatives; this is crucial for medical imaging applications.

Figure 11 describes the trends of specificity of the model representing its ability to correctly classify negative cases. The steady and high validation specificity suggests that the model well distinguishes the normal X-rays from those affected with cardiomegaly and reduces the number of false positives, increasing diagnostic precision.

The model sensitivity indicating how well the model identifies true positive cases is shown by the trends of these metrics in Figure 12. The validation sensitivity is about, above 90%, which is actually a good sign of being able to detect cardiomegaly without missing significant cases; hence, it assures being reliable for automated diagnosis.

In Figure 13, the AUC values of the training and validation concerning the model's distinction of affected and unaffected cases were shown. The AUC hung around equally high scores, indicating good discriminative power, thus further reaffirming the robustness of the model in clinical use.

**Table 3. Accuracy Comparison across Models**

| Papers | Accuracies |
|---|---|
| **[6] Iqbal et al**. | 92.0% |
| **[3] Bar et al.** | 91.3% |
| **[8] Rubin et al.** | 89.0% |
| **[4] Bar et al.** | 92.5% |
| **Proposed [CMMCA-V3]** | 95.6% |

Table 3 depicts the comparison of the accuracy of the proposed model with the older studies. The proposed method has achieved the highest accuracy so far (95.6%) and has surpassed existing models by incorporating multi-head attention with InceptionV3. This shows the suggested method's efficiency in improving classification performance.

**Comparative Analysis**

The following table showcases the accuracies from previously published papers with our proposed algorithm that uses Multi head Attention Enhanced InceptionV3. We can infer that our model performs significantly better than [26] which uses heavy models such as RestNet-101. It also performed better than [27] that used a large MIMIC-CXR dataset.

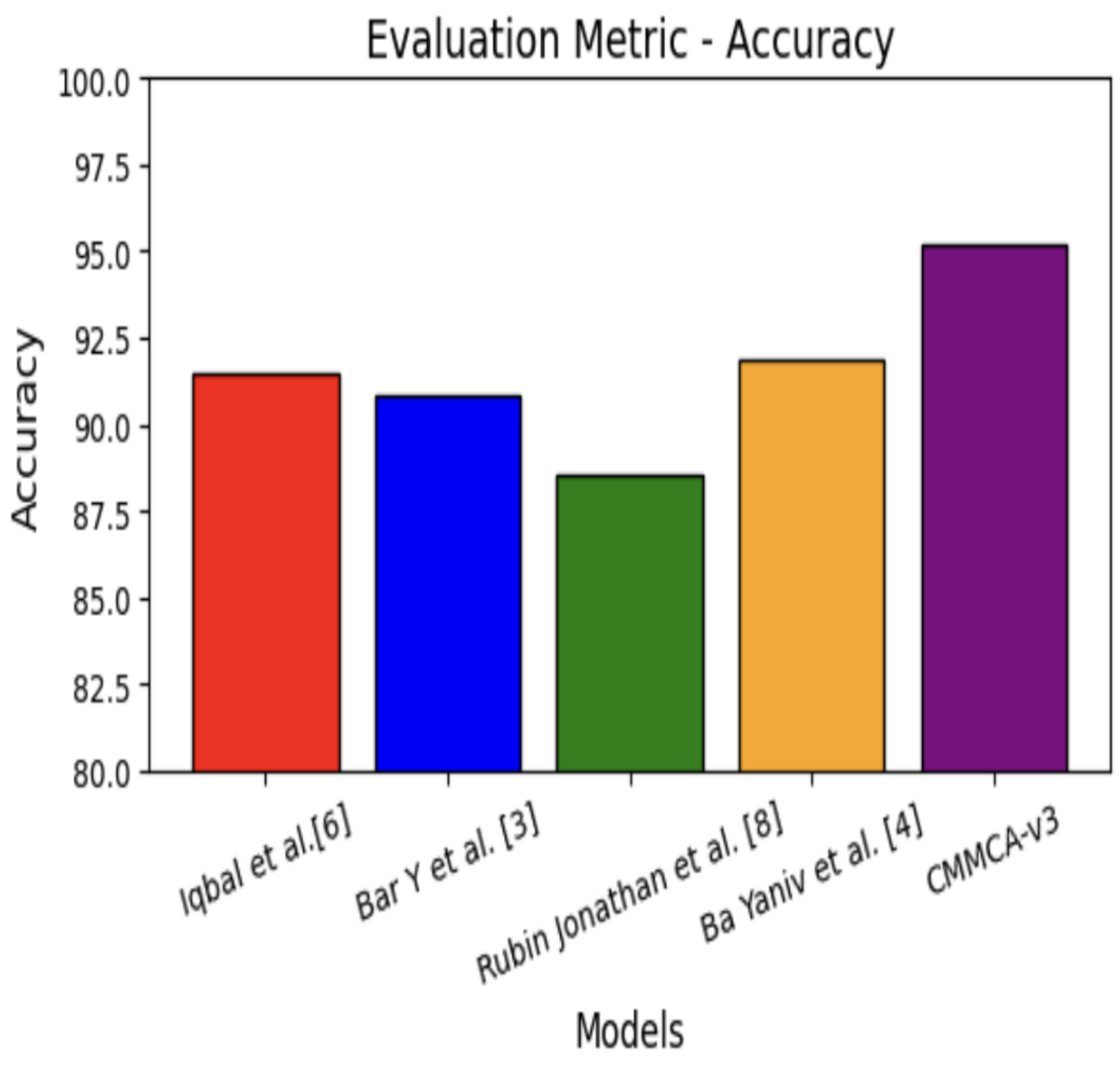

Figure 14. Accuracy Comparison

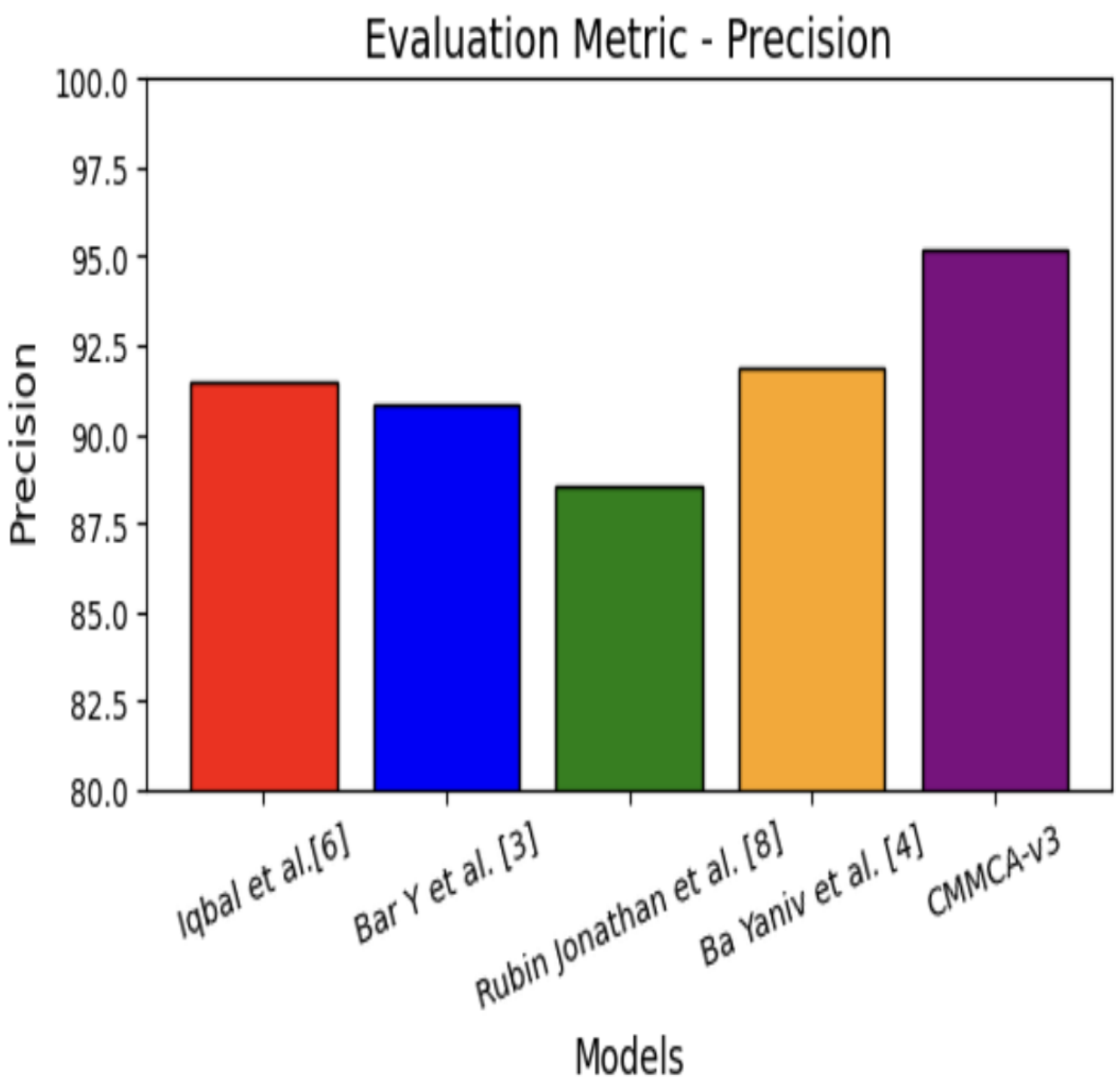

Figure 15. Precision Comparison

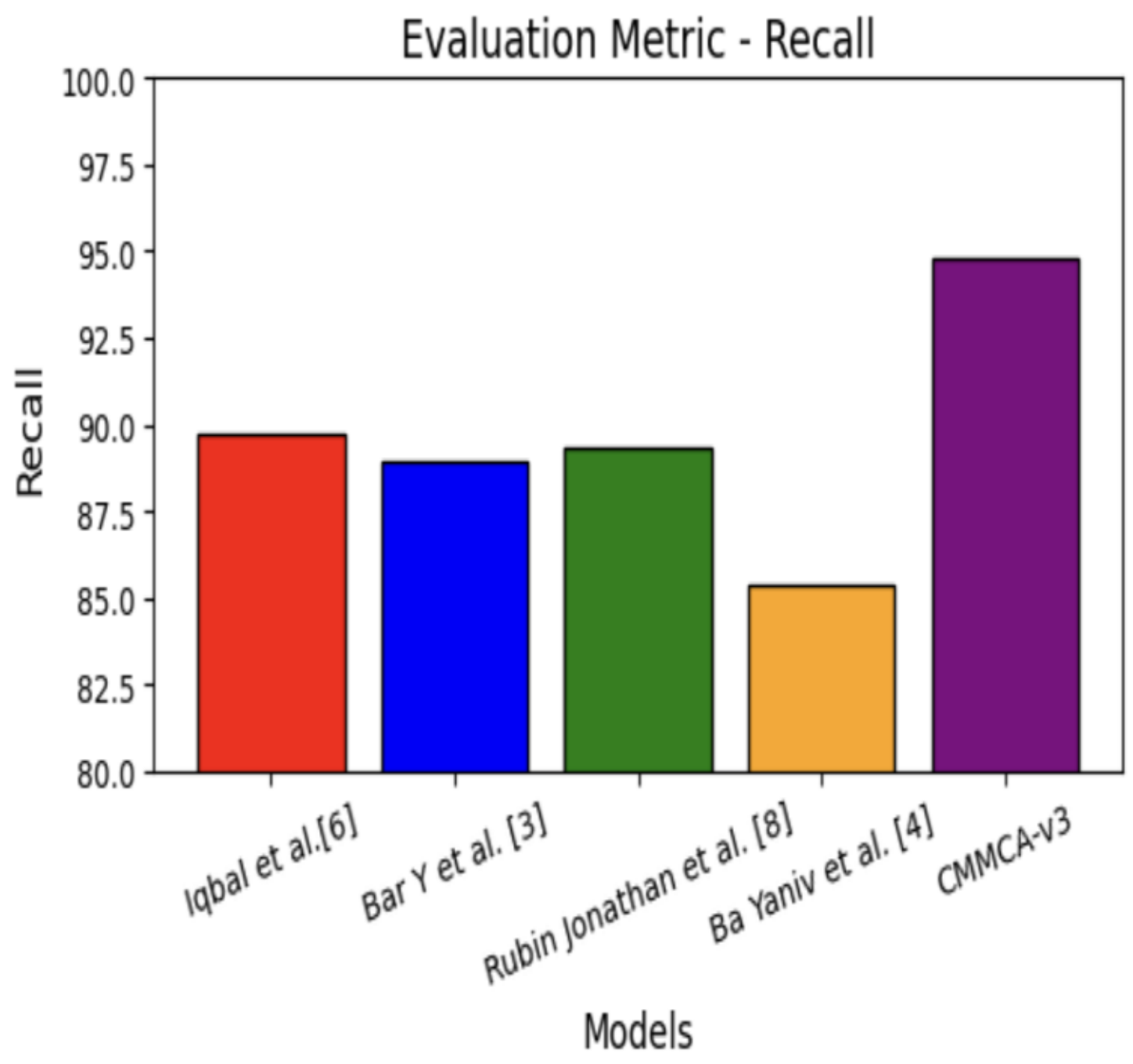

**Figure 16. Recall Comparison**

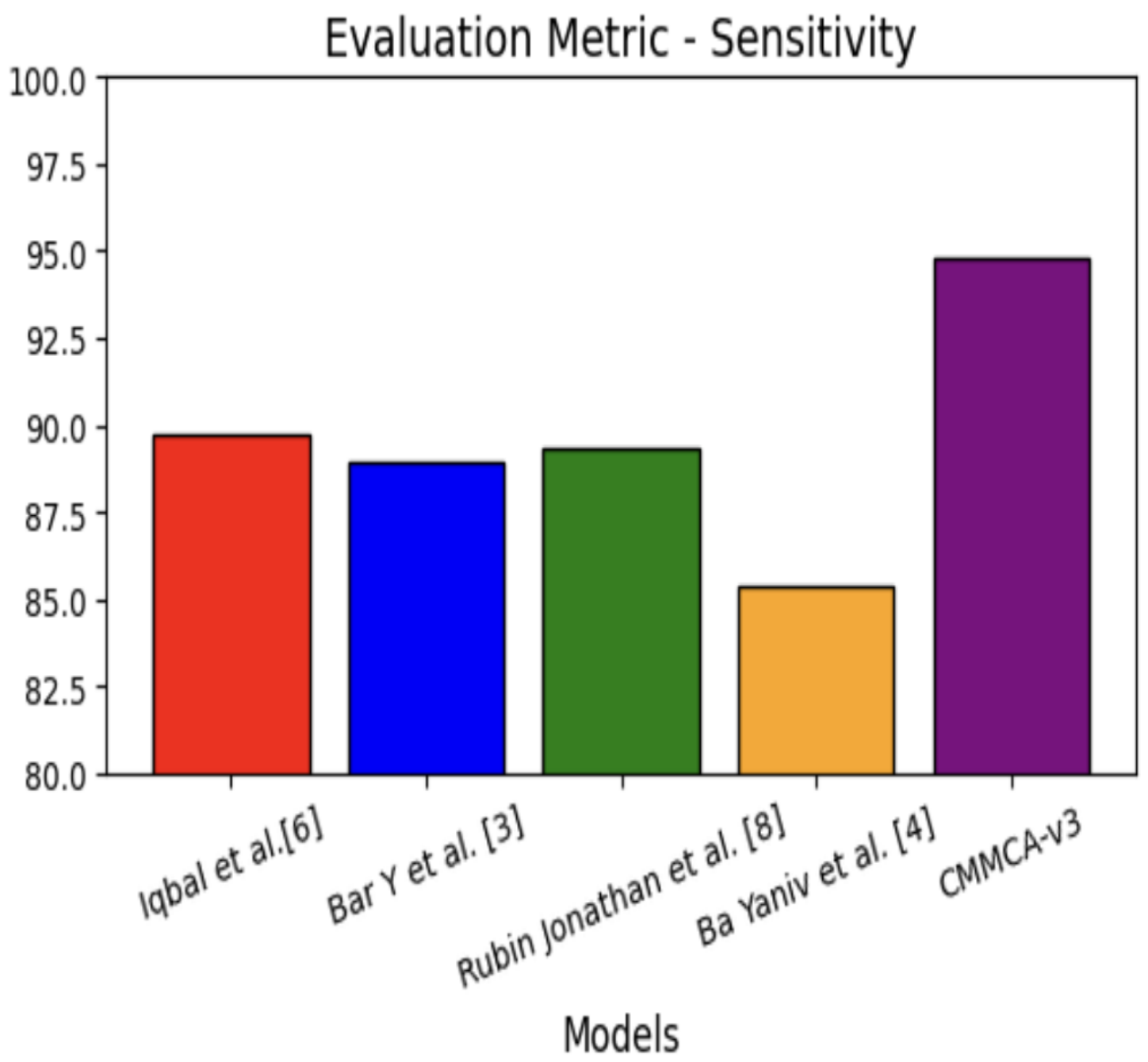

**Figure 17. Sensitivity Comparison**

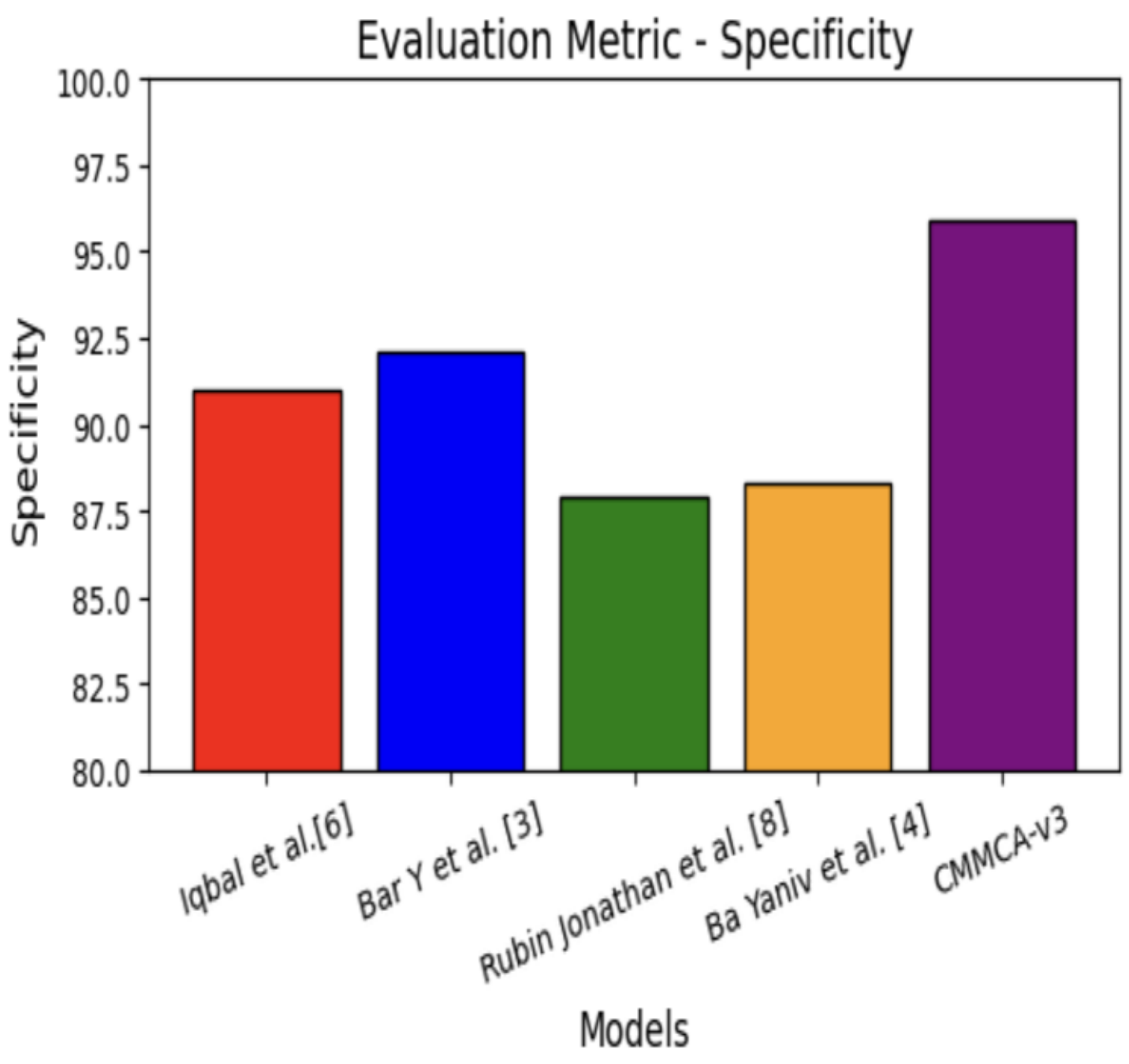

**Figure 18. Specificity Comparison**

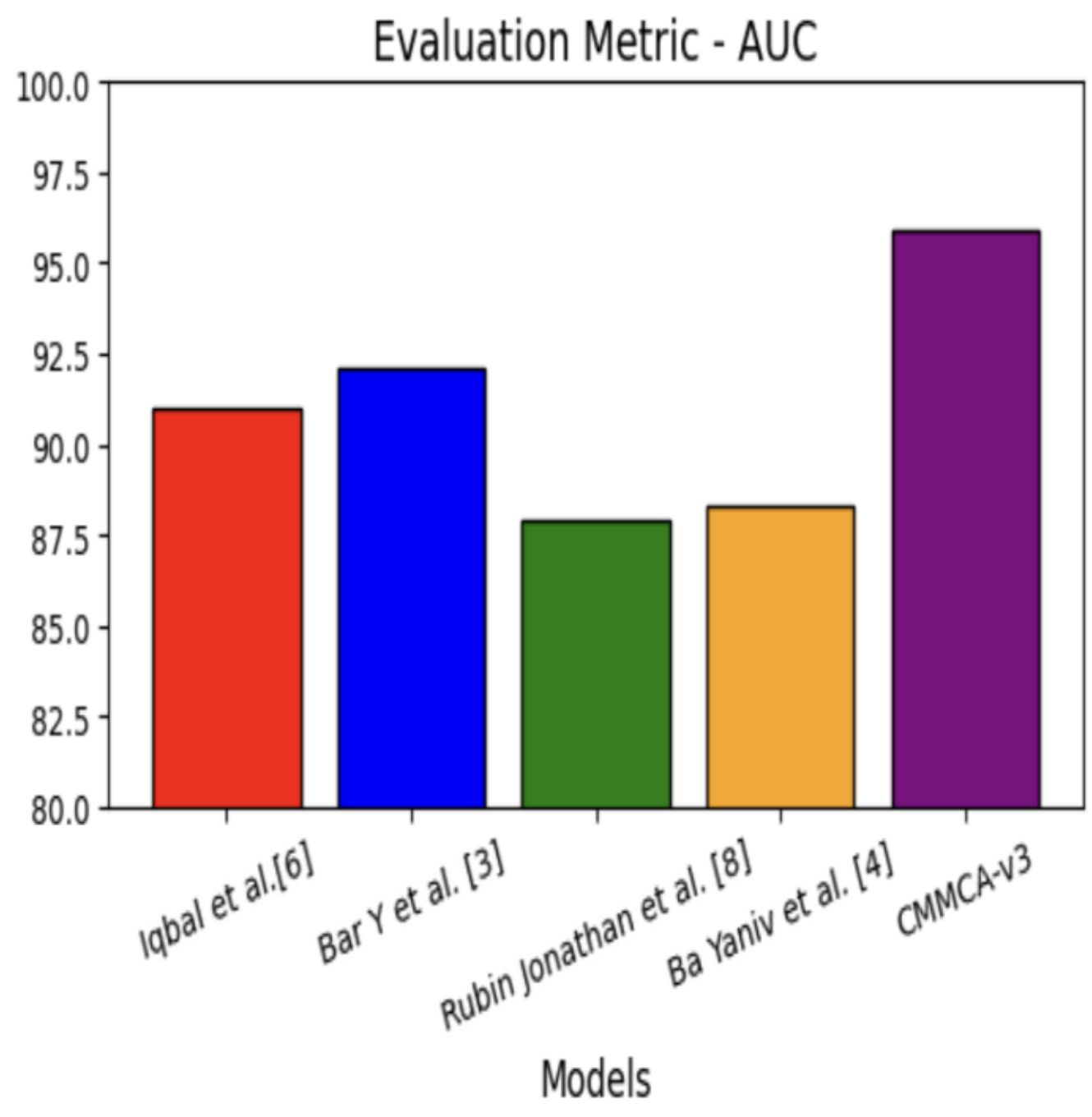

**Figure 19. AUC Comparison**

A comparative analysis of the accuracy of the proposed model and past studies is shown in figure 14. The highest accuracy results show that the new model, empowered by multi-head attention with InceptionV3, surpasses all prior models. Its improvement in accuracy indicates the usefulness of incorporating attention mechanisms to extract better features for chest X-ray analysis.

Performance comparison of precision between the proposed model and past methods is shown in figure 15. Precision indicates the fraction of the correctly predicted positive cases upon the total predicted positive cases. The proposed model achieves better precision, resulting in fewer false positives and, therefore, increased reliability for cardiomegaly detection.

Figure 16 presents a comparison of different models based on recall, demonstrating how well each model could identify cardiomegaly cases with the proposed model performing better than previous models with a recall value far higher. This will give a larger number of patients suffering from cardiomegaly the possibility to be finally diagnosed correctly and will thus reduce the possibility for undergoing false-negative results.

Figure 17 presents the sensitivity comparison showing that proposed method is more sensitive than other techniques. It is reasonably more trustworthy if the sensitivity is more, as it would have fewer numbers of undiagnosed cases from the index model, which is especially required in clinical environments where missing any case may require serious attention.

The Figure 18 displays the specificity comparison with different models. Proposed model yields higher specificity indicating better performance to reduce false positives. This is of utmost importance to the medical field as any misclassification may lead to unnecessary treatment or further tests.

The AUC Comparison in Figure 19 shows the comparison of AUC values with different models. The proposed model achieves the highest AUC indicating better performance of the classifications, signifying great confidence in the model that can distinguish between normal and cardiomegaly cases, thus improving its clinical applicability.

## Conclusion

Ultimately, the creation of a comprehensive technology that is autonomous in diagnosing cardiomegaly in X-ray images is a significant milestone in medical imaging. The integration of imaging technology, deep learning techniques, and attention mechanisms has made possible the implementation of an automated solution that is accurate and efficient in cardiomegaly diagnosis. Proposed method, with data collection, preprocessing, model design, and testing, has shown promising outcomes in the detection of cardiomegaly in thoracic radiographs. Using the convolutional neural network (CNN) structure with InceptionV3 as the primary one, applying multi-head attention mechanism to it, the system is able to achieve high precision and recall that are the most important measures for the system being accurate in its diagnosis and patient care. In the same way, the performance measures assessing accuracy, precision, recall, specificity, sensitivity and F1 score provide an overall picture of the model's capability and shortcomings. The conclusion is that the analyst is able to know how the model performs by going through graphical illustrations like training and validation metrics and ROC curve will indicate how the model converges and generalizes, whereas the analysis of confusion matrix will provide some insight into potential areas where the model can be refined, specifically minimizing false positives and false negatives.

## Future Work

Future work can enhance data quality and diagnostic accuracy by leveraging advanced preprocessing techniques such as image augmentation while extracting domain-relevant features. Multi-modal data, including clinical notes and demographic details, further contributes to performance improvement. Significant progress will be made in medical imaging, while investments in automated imaging analysis may be promising for the future of health. The optimization of the proposed method along with advanced technologies is likely to enhance the diagnostic accuracy and enable great advances in radiology. Such domains must call for the attention of future researchers for such advancements in medical AI applications.